\documentclass[11pt,a4paper]{article}
\usepackage[hyperref]{emnlp2020}
\usepackage{times}
\usepackage{latexsym}

\usepackage{booktabs}
\usepackage{subcaption}
\usepackage{dcolumn}
\usepackage{microtype}
\usepackage{paralist}
\usepackage{float}
\aclfinalcopy 

\usepackage{amsmath}
\usepackage{amssymb}
\usepackage{graphicx}
\usepackage{enumitem}
\setitemize{itemsep=0.5pt}

\title{With Little Power Comes Great Responsibility}

\author{Dallas Card$^1$ \quad
  Peter Henderson$^1$ \quad
  Urvashi Khandelwal$^1$ \quad
  Robin Jia$^1$  \\
  \textbf{Kyle Mahowald}$^2$ \quad
  \textbf{Dan Jurafsky}$^1$ \\
  $^1$Stanford University, Stanford, CA\\
  $^2$University of California Santa Barbara, Santa Barbara, CA\\
  \texttt{dcard@stanford.edu}, \texttt{phend@stanford.edu}, \\ \texttt{urvashik@stanford.edu}, \texttt{robinjia@stanford.edu}, \\ \texttt{mahowald@ucsb.edu}, \texttt{jurafsky@stanford.edu}}
\date{}

\newif\ifcomment
\commenttrue

\ifcomment
\newcommand{\uk}[1]{\textcolor{red}{UK: #1}}
\newcommand{\dc}[1]{\textcolor{blue}{DC: #1}}
\newcommand{\kyle}[1]{\textcolor{orange}{KM: #1}}
\newcommand{\ph}[1]{\textcolor{cyan}{PH: #1}}
\newcommand{\dan}[1]{\textcolor{magenta}{DJ: #1}}
\newcommand{\rj}[1]{\textcolor{magenta}{RJ: #1}}
\else
\newcommand{\uk}[1]{}
\newcommand{\dc}[1]{}
\newcommand{\kyle}[1]{}
\newcommand{\dan}[1]{}
\newcommand{\ph}[1]{}
\newcommand{\rj}[1]{}
\fi

\begin{document}
\maketitle
\begin{abstract}
Despite its importance to experimental design, statistical power (the probability that, given a real effect, an experiment will reject the null hypothesis) has largely been ignored by the NLP community.
Underpowered experiments make it more difficult to discern the difference between statistical noise and meaningful model improvements, and increase the chances of exaggerated findings. 
By meta-analyzing a set of existing NLP papers and datasets, we characterize typical power for a variety of settings and conclude that underpowered experiments are common in the NLP literature.
In particular, for
several tasks in the popular GLUE benchmark, 
small test sets mean that most attempted comparisons to state of the art models will not be adequately powered.
Similarly, based on reasonable assumptions, we find that the most typical experimental design for human rating studies will be  underpowered to detect small model differences, of the sort that are frequently studied.
For machine translation, we find that typical test sets of 2000 sentences have approximately 75\% power
to detect differences of 
1 BLEU point.
To improve the situation going forward, we give an overview of best practices for power analysis in NLP and release a series of notebooks
to assist with future power analyses.\footnote{\href{https://github.com/dallascard/NLP-power-analysis}{https://github.com/dallascard/NLP-power-analysis}}
\end{abstract}

\section{Introduction}

Despite its importance to empirical evaluation, relatively little attention has been paid to statistical power in NLP.
In particular, \emph{if it is the case that typical experiments in NLP are underpowered}, not only would we expect many meaningful improvements to go undetected, we would also expect many apparently significant differences to be exaggerated \citep{gelman2014beyond}.
In this paper, we build on past work calling for greater rigor in evaluation \citep{mccoy2019berts,azer.2020}, including the need for careful hypothesis testing \cite{koehn.2004,berg.2012,sogaard.2014,dror.2018}, 
and show why and how power matters to NLP,
addressing challenges unique to this domain.

\begin{figure}
    \centering
    \includegraphics[scale=0.55]{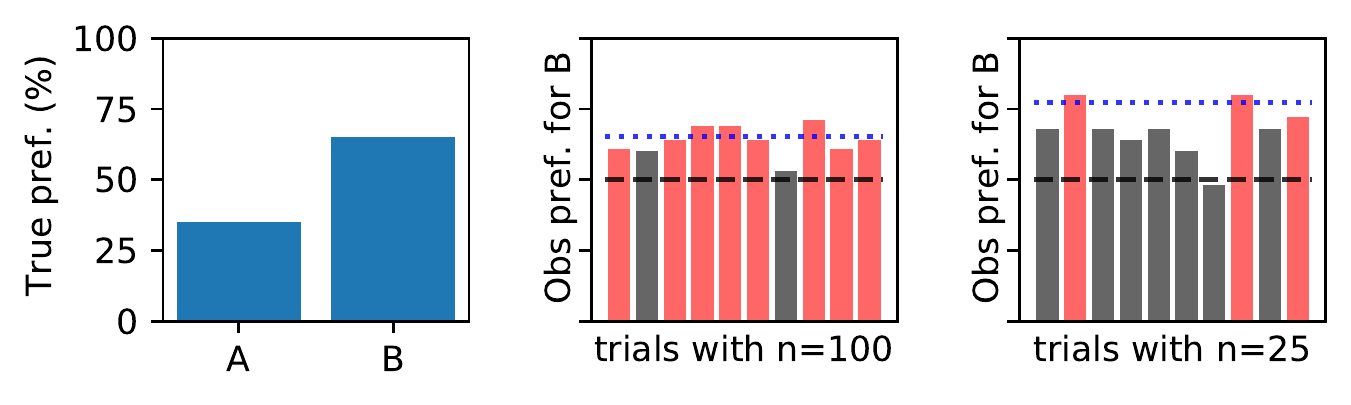}
    \caption{Cartoon example of statistical power in comparing two models: 65\% of all people in the population always prefer system B (left). A comparison using a sample of $100$ people would be well-powered (middle): over 80\% of such samples will show a significant difference (plotted in red) from the null hypothesis that the models are equally good (dashed line). In samples of $25$ people (right), far fewer tests will be significant (power $\approx30\%$). Note that the observed mean of significant findings (dotted line) slightly overestimates the true proportion that prefer system B when $n=100$ and more severely overestimates it when $n=25$. }
    \label{fig:ABdemo}
\end{figure}

Roughly speaking, power is the probability that a statistical test will successfully detect a \emph{true effect}.
As an illustrative example, imagine 
comparing two dialog systems
(see Figure \ref{fig:ABdemo}). We want to know if people tend to prefer one system over the other. To test this, we will need multiple people to evaluate the systems. But how many?
Once we have collected data, a statistical \emph{test} will tell us if we can reject the null hypothesis the systems are equally good.
Assuming the systems are not identical, 
statistical \emph{power} is the probability that the experiment
will return a significant result
(or equivalently, it is one minus the probability of failing to detect the difference as significant).
Although we don't know the 
magnitude of this difference,
\emph{power analysis} helps to estimate how much power an experiment will have under various assumptions. 

Power depends on multiple factors, including the statistical test used,
the
significance threshold,
true effect size,
variance, and sample size. All else being equal, experiments with larger samples will have greater power than smaller samples, as shown in Figure \ref{fig:ABdemo}.
Similarly, larger effects and those with less variance are easier to detect, and therefore require fewer samples for equivalent power.
Importantly, note that if we \emph{do} find a significant difference, this does \emph{not} imply that the experiment had high power.\footnote{Using the observed outcome from a single experiment to compute power falls into the trap of post-hoc power analysis and is not recommended. For additional background on statistical power, power analysis, null-hypothesis significance testing, and post-hoc analysis, please refer to Appendix~\ref{app:nhst}. }

Proceeding with a test that is \emph{underpowered} (i.e., too few subjects or items; often taken to mean less than 80\% power; \citealp{cohen_statistical_1962}) means that one is less likely to be able to draw any useful statistical conclusion from the experiment,
and has contributed, in part, to the replication crisis in other fields \citep{button2013power,szucs2017empirical,ioannidis.2017}. 
Routinely
running experiments with low statistical power
undermines the scientific enterprise.
Not only
will true effects go undetected;
when
significant effects
are found,
they are likely to
be noisier
and 
have lower positive predictive value \citep{button2013power}.

Moreover, significant findings from underpowered experiments 
are more likely to exaggerate or reverse the true effect -- so-called Type-M (magnitude) and Type-S (sign) errors, respectively \citep{gelman2014beyond}.
This problem can lead to systematic distortions in the literature if only significant findings are published, especially if these results are based on underpowered experiments \citep{scargle.1999}. The effect of Type-M error can be seen in Figure \ref{fig:ABdemo}; significant differences are less likely to be found in smaller samples (right), but among those tests that are significant, the observed difference will tend to exaggerate the true difference (left) by more than a larger sample (middle). For further discussion of Type-M and Type-S errors, please refer to Appendix \ref{app:typem}.

Here, we investigate how these issues affect NLP. 
Although retrospective analysis of power involves challenges, we present evidence that underpowered experiments are widespread in NLP research. Among human evaluations, we find most experimental designs involve too few items and/or raters to detect small effects (\S\ref{sec:humaneval}). For comparing models in terms of accuracy, we find that some widely used benchmark datasets, including MRPC and SST-2, are now too small to be able to properly measure future progress against top performing models (\S\ref{sec:accuracy}). We also introduce a novel approach to power analysis for machine translation and characterize power in experiments testing for differences in BLEU (\S\ref{sec:mt}).
Finally, a survey of recent papers reveals a general lack of statistical evaluation and a dearth of detailed reporting (\S\ref{sec:meta}).

To improve future practice, we suggest broader adoption of power analyses prior to evaluation, provide guidance on running power analyses in NLP, and
release a series of notebooks for this purpose. 

\section{Power Analysis for NLP}
\label{sec:nlp_power}

Because most NLP tasks do not take the form of standard experiments in other sciences \citep{kraemer.2015,westfall.2014}, it is non-trivial to run power analyses for many tasks of interest.
While we cannot cover every scenario, we present here a generalizable, simulation-based approach to power analysis, along with three sample applications, which can be extended as necessary.
Such an approach is modular, reusable, and transparent, and encourages planning of analyses in advance of data collection.

Every power analysis requires assumptions, and there is not likely to be a single correct approach. Rather, the point is to make one's assumptions explicit, and include enough detail so as to account for whatever is likely to be observed. By using reasonable assumptions, one can help to ensure that one's experiment is sufficiently well-powered, In the case of NLP, this means that one recruits enough subjects, collects enough ratings, or uses a large enough test set.

The general procedure we suggest for power analysis is described in detail in Figure \ref{fig:algo}. At a high level, the idea is to estimate power by running simulations. Recall that power is the probability of detecting a true effect, conditional on the experimental setting (effect size, variance, etc.) and significance threshold. Thus, if one can translate these assumptions into a process for generating simulated data, we can estimate power by generating many simulated datasets using assumed or estimated parameter values, running each sample through a significance test, and reporting the proportion that are found to be significant.

\begin{figure}
\centering \small
\fbox{\parbox{0.47\textwidth}{
Define a generative process $G(n, e^*, \mathbf{h})$ parameterized by number of items, $n$, hypothesized effect $e^*$ for the statistic of interest $E$, and other relevant parameters $\mathbf{h}$ (e.g., variance). Also choose a statistical test $T(\mathcal{D})$, which returns a p-value $p$ when performed on data $\mathcal{D}$ sampled from $G(n, e^*, \mathbf{h})$. Finally, choose the size of the dataset to be sampled, $n$, significance threshold, $\alpha$, and number of repetitions, $r$.
\begin{enumerate}
\item For $i$ in range($r$):
 \begin{itemize}
 \item sample a dataset of size $n$, $\mathcal{D}_i \sim G(n, e^*, \mathbf{h})$
 \item compute the effect of interest on this sample, $e_i = E(\mathcal{D}_i)$
 \item also compute a p-value according to the test of interest: $p_i = T(D_i)$
 \end{itemize}
\item $\textrm{Power} \approx \frac{1}{r} \sum (\mathbb{I}[p_i \leq \alpha] \cdot \mathbb{I}[\textrm{sign}(e_i)=\textrm{sign}(e^*)])$
\end{enumerate}
} }
\caption{An algorithm for power analysis by simulation. For the example of comparing two systems presented in Figure \ref{fig:ABdemo}, $e^*$ is the assumed overall proportion of people who prefer system B, relative to the null hypothesis, $p=0.5$, $G(n, e^*, \mathbf{h})$ is simply $\textrm{Binomial}(n, 0.5+e^*)$, while $e_i$ is the observed proportion of people who prefer system B in sample $i$, again relative to 0.5.
For extensions to estimate Type-M and Type-S error, see Appendix \ref{app:typem}.}
\label{fig:algo}
\end{figure}

The key to generalizing this approach is to begin with the end in mind. In particular, if one plans to test for a difference between models, one needs to choose the statistical test that will be used. That test will determine the level of detail required in the generative process for simulating data.

To return to the opening example of evaluating dialog systems,  we want to test if people prefer one system over the other \citep{ai.2007}. If we ignore the nuances of human preference for now (but see \S\ref{sec:humaneval} for a more nuanced approach), and simply assume that each person either prefers system A or system B, the only assumption we need to make for a power analysis in this setting is the proportion of people in the population who prefer system B. We can then simulate samples of $n$ people (each of whom independently has the same probability of preferring system B) as a draw from a binomial distribution, and repeat this thousands of times.\footnote{We don't need to address variance
in this scenario, as the variance of a binomial distribution is a function of its mean.} For each sample, we then test whether the proportion of people who prefer system B is significantly different from 0.5. The estimated power of this experiment would thus be the proportion of simulated differences that are found to be significant.\footnote{More direct solutions are available for some settings, including this one (see Appendix \ref{app:binomial}), but we describe it using the generic approach from Figure \ref{fig:algo} for the purpose of illustration. For all cases examined in this paper, simulations take only minutes on a laptop.}

The most difficult part of power analyses is estimating the relevant quantities, such as the \emph{true} proportion of people that prefer system B. Note, however, that one can always compute what power would be for a range of possible values, and indeed, this is the recommended procedure.
For estimating the relevant parameters within an NLP context, we will primarily rely on data from the literature, measurements on validation data, and estimates from external datasets (see \S\ref{sec:acc_params}).
However, where appropriate, 
pilot studies may also be informative.

In the remainder of this paper, we 
consider
three scenarios of interest
in depth, 
and assess the state of power in the NLP literature for each.

\section{Comparing Models on Accuracy} \label{sec:accuracy}
It is common in 
NLP research to look for models which improve over state of the art (SOTA) on various benchmarks.
However, an important but rarely asked question is, \emph{can these benchmarks support the kinds of comparisons we want to make?}
Many have emphasized the need for proper significance testing to avoid spurious findings, but 
if an experiment's test set is small, the minimum detectable effect (MDE) size 
may be large:
only large 
improvements will 
yield sufficiently powered comparisons (i.e., $\geq 80\%$ power).
If an experiment is badly underpowered, it cannot provide useful evidence that one model achieves slightly better performance than another for the underlying data distribution. Reliance on such evidence risks leading to over-confidence about the relative ranking of various models.
As we show in \S\ref{sec:acc_lit}, there is legitimate reason to be concerned about this in the case of certain widely used benchmarks.

\subsection{Significance test for comparing classifiers}
\label{sec:mcnemar}

The standard statistical test for comparing classifiers on paired data is McNemar's test \citep{dietterich1998approximate,dror.2018}, which uses the numbers of items where the models disagree (i.e., the off-diagonal elements in Table \ref{tab:contingency}).\footnote{Unpaired data (i.e., if two models are evaluated on different data drawn from the same distribution) requires a different approach, such as using a binomial test. See Appendix~\ref{app:binomial} for extended discussion.} 
McNemar's test assesses whether $\chi^2 = \frac{\left( p_{10} - p_{01} \right)^2}{p_{10} + p_{01}}$ is significant, and if so, rejects the null hypothesis that the distributions are the same.  
\begin{table}[!htb]
\begin{center}
\small
\begin{tabular}{l|cc}
        & M1 correct & M1 incorrect\\\hline
M2 correct & $\textrm{both correct}$ & $\textrm{only M2 correct}$\\ 
M2 incorrect & $\textrm{only M1 correct}$ & $\textrm{both incorrect}$\\ 
\end{tabular}
\end{center}
    \caption{A contingency table representing the distribution of possible outcomes for two models (M1 and M2).}
    \label{tab:contingency}
\end{table}

\begin{figure}[ht]
    \centering
    \includegraphics[scale=0.53]{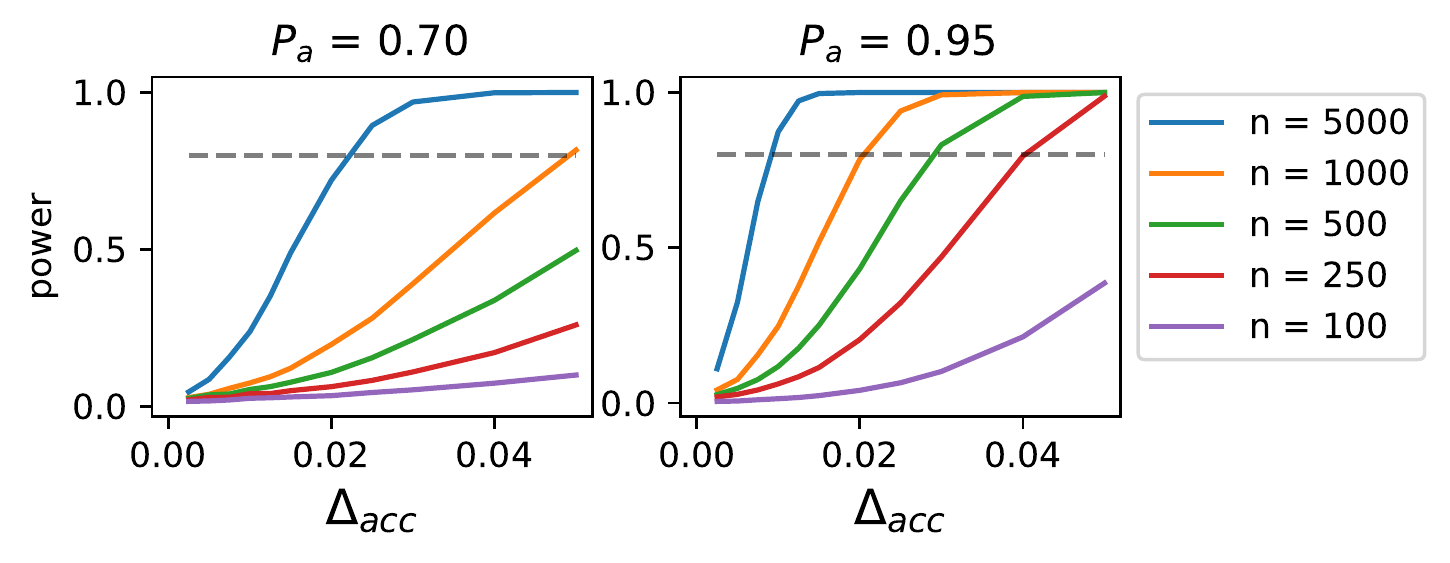}
    \caption{
    Power for comparing two classifiers 
    on accuracy using paired data 
    depends on
    the size of the test set ($n$), the expected agreement ($P_a$), and the expected difference in accuracy ($\Delta_{acc}$). 
    The 
    dashed 
    line shows 80\% power, 
    often taken to be
    a
    minimal requirement.
    }
    \label{fig:ref_acc}
\end{figure}

Thus, for McNemar's test, the relevant data generating process for simulations can be specified in terms of the expected difference in accuracy between the models, $\Delta_{acc}$, and $P_a$, the expected proportion of examples for which the models will have the same outcome (i.e., both correct or both incorrect).
From these
we can compute the expected proportions of examples on which only one model is correct (i.e., the off-diagonals in Table \ref{tab:contingency}), and estimate power via the algorithm in Figure~\ref{fig:algo}.
Figure \ref{fig:ref_acc} 
illustrates how power increases with increased sample size, effect size, and agreement rate.\footnote{Corresponding plots showing Type-M and Type-S error \citep{gelman2014beyond} are in Appendix \ref{app:typem}.
To walk through a 
numerical example, see Appendix~\ref{app:mcnemar_example}. For an interactive example, see the accompanying online notebooks. }

\subsection{Estimating parameters}
\label{sec:acc_params}

In order to estimate the required parameters ($P_a$ and $\Delta_{acc}$), we consider three options:
(1) use results on validation (dev) data; (2) fit a regression
based
on 
historical data; (3) use middle-of-the-road assumptions when lacking other information.
Using these methods, we can then estimate power or calculate the smallest effect that can be detected with 80\% power at $\alpha=0.05$ (or other thresholds).
Both to illustrate this process, and to provide guidance for future work, we demonstrate these approaches below using data from two widely-used datasets for evaluating NLP models: SQuAD 2.0~\cite{rajpurkar2016squad,rajpurkar2018know} and the GLUE  benchmark \citep{wang2018glue}.

\paragraph{Using validation results:} 
To the extent that we expect performance on test data to match performance on validation data (i.e., in the absence of domain shift), 
\emph{paired} performance on validation data (i.e., difference in accuracy and agreement rate) provides one method for estimating power 
when
comparing against a baseline model.

To illustrate this, from the authors of SQuAD 2.0, we obtain the pairwise agreement rates between all models submitted to the leaderboard on both validation and test data.
We find
a very strong correlation between validation and test for both pairwise accuracy differences ($\Delta_{acc})$ and agreement rates ($P_a$)
($r=0.99$ for both, as shown in Figure \ref{fig:squad_correlation} in Appendix \ref{app:squad}, with results on validation data included in the accompanying online materials), suggesting we can use paired predictions on validation data for power calculations when we have access to the predictions from both models.
Note that this approach assumes that the dev and test data have been drawn from the same distribution, and that dev performance has not been artificially inflated (such as by training on validation data directly).

\paragraph{Using historical data:}
When one does not have access to the baseline model
or an informative prior,
one can make use of historical trends. That is, we can try to estimate what a typical improvement will look like, given the current state of the art (SOTA).
To illustrate this approach, we collect reported results for both SQuAD 2.0 and GLUE, and fit regressions to estimate $\Delta_{acc}$ and $P_a$. Given these parameters, we can assess the likely power and MDE for a typical model improvement against a given baseline accuracy level.

To fit a
regression 
to predict
typical improvements to SOTA,
we gather data from GLUE papers and manually label 119 accuracy comparisons and 57 claims of improvement (as denoted
by bolding of a
result
and a claim of SOTA
in text) across 14 papers (selected as being at or above the BERT score on the GLUE leaderboard with an accompanying paper). 
In regressing $\Delta_{acc}$ on baseline accuracy and task,
we achieve an $R^2 = 0.69$, which is not a perfect fit, but still provides a prior on likely effect size. Similarly, we achieve an $R^2=0.67$ when fitting a regression to SOTA improvements on the SQuAD 2.0 leaderboard (selected as being a significant improvement in time-ordered submissions). See Appendix~\ref{app:overlap} for more details.

To assess power for McNemar's test, we must also fit a regression predicting the expected overlap between the models ($P_a$). To fit such a regression, from
GLUE authors
we obtain the model test set predictions on all tasks from a set of 10 high-performing models, which allows us to measure the extent to which their predictions overlap with each other. 
Using GLUE tasks which measure accuracy,
we regress $P_a$ on baseline accuracy and $\Delta_{acc}$, and 
achieve an $R^2$ of $0.97$.\footnote{
WNLI~\citep{levesque2012winograd},
MRPC~\citep{dolan-brockett-2005-automatically},
SST-2~\citep{socher2013recursive},
RTE~\citep{dagan2005pascal,bar2006second,giampiccolo2007third,bentivogli2009fifth},
QNLI~\citep{rajpurkar2016squad}
MNLI~\citep{williams2018broad}, and QQP~\citep{iyer2017first}. For consideration of other metrics, see Appendix \ref{app:metrics}.  } 
Repeating this for SQuAD 2.0,
we get an $R^2$ of $0.94$. See Appendix~\ref{app:regressions} for regression coefficients and additional details.

Typical improvements on popular tasks tend to be small (see mean improvements in Table~\ref{tab:glue}). Except for rare transformative work, such as BERT \citep{devlin2018bert}, it is generally difficult to do \emph{much} better than a previous SOTA and thus improvements are likely to follow a trend, which is why we are able to use historical
data as a guide.
In cases where such data is not available or cannot be trusted, other methods are necessary.

\paragraph{No prior:} If no informative prior 
is available 
and the baseline model or can't be used for comparison on a validation set, 
then we must fall back on middle of the road assumptions.
\citet{lachenbruch1992sample} provides a suggested default prior, and we find that MDEs using this method are very similar to those found by using the regression based approach.  Appendix~\ref{app:lachenbruch} provides more details, and Table \ref{tab:glueexpanded} in the appendix presents the comparison.

\subsection{Assessing power in the literature}
\label{sec:acc_lit}

Using the regression-based approach of estimating $\Delta_{acc}$ and $P_a$ described above, we estimate the MDE for each individual accuracy-based GLUE task in comparison to current SOTA, and report the average effect size of results which claimed improvements.
Table \ref{tab:glue} summarizes these results,
showing for each dataset the size of the test set, 
the accuracy of the best performing model on each task at the time of
writing,
the estimated MDE to have 80\% power using our regression 
to predict overlap ($P_a$),
and 
the average reported difference from 
their respective baselines. 

\begin{table}[]
    \centering
    \resizebox{.48\textwidth}{!}{
    \begin{tabular}{lcccc}
         \toprule
         \textbf{Dataset} & \textbf{Size} & \textbf{SOTA} (\%) &  \textbf{Est. MDE} (\%) & $|\Delta_{acc}|$ (\%)\\
         \midrule
         WNLI&147&94.5&+5.26 & +1.72 \\
      MRPC&1,725& 92.0  & +1.62 & +0.63\\      
      SST-2& 1,821 & 97.2 &  +1.02 & +0.57 \\
      RTE&3,000 & 91.7& +1.23 &  +3.89 \\      
      QNLI&5,463&97.5& +0.55 &  +1.31 \\
      MNLI-m & 9,796 & 91.6 & +0.67 & +0.97 \\
      MNLI-mm & 9,847& 91.3& +0.68 &  +1.29 \\
      QQP &390,965&91.0&  +0.11 &  +0.36 \\
    \midrule
    SQuAD 2.0 & 8,862&90.7 & +0.56& +2.23$^\dagger$\\   
      \bottomrule
    \end{tabular}
    }
    \caption{Estimated minimum detectable effect (MDE) using a regression-based estimate of likely agreement with leaderboard SOTA as of May 6th, 2020.
    $|\Delta_{acc}|$ is the average improvement over baseline per task among surveyed papers that claimed SOTA.
    For future comparisons, unless the expected improvement is larger than the estimated MDE, an experiment is unlikely to be adequately powered, and researchers should instead choose a different (larger) dataset.
    Note that this likely applies to the vast majority of experiments on WNLI, MRPC, and SST-2, based on recent trends.
     $\dagger$ indicates that the SQuAD 2.0 average was based on leaderboard improvements, which weren't necessarily reported in a publication. 
     See Appendix~\ref{app:mde_table} for full table and details. }
    \label{tab:glue}
\end{table}

As can be seen
in Table \ref{tab:glue},
the mean reported effect size ($|\Delta_{acc}|$)
is well below the 
estimated MDE for the three smallest test sets -- WNLI, MRPC, and SST-2.
Because this mean is based on models comparing to even weaker baselines, we would expect most future improvements to be even smaller.
Thus, most future experiments involving these three datasets \emph{will not have adequate power} to test for improvements over the current SOTA in the way that they are routinely used.
Moreover, alternative analyses give \emph{even more pessimistic} estimates of likely improvements relative to MDE, 
as described in Appendix \ref{app:glueexpandedres}.
If an experiment does show significant improvement on a dataset such as MRPC, the potential for Type-M error should make us skeptical that this improvement will generalize to new data from the same domain.

While the above results are informative about future experiments, 
we would
also
ideally 
like to know about the power of past experiments.
Most of the papers from which we collected results did not report a significance test on the test set. Here we estimate the expected power and predicted result of such a test using leave-one-out regressions, where we make a prediction for each reported improvement using all other reported model comparisons. This procedure reveals that \textbf{only 46\% would have predicted adequate power} (using estimates for expected improvement and agreement), and \textbf{approximately 51\% would have been significant} (based on estimated agreement and \emph{reported} improvement). Approximately 80\% of experiments with at least 80\% power would also have been found to be significant (37\% of all comparisons).

In part because performance on many of these tasks is now so good, a large expected improvement is required in order for 
a new experiment 
to have 80\% power, suggesting that larger test set sizes may be necessary to continue making well-powered claims of SOTA improvement on individual tasks.
For any comparisons which are likely to be underpowered, we should refrain from placing much emphasis on obtaining small improvements over the previously reported best model. In extreme cases, such as MRPC and SST-2, it is worth considering whether it is time to retire these datasets as the basis for model comparison.\footnote{
It is also worth exploring power with respect to claims of improvement on multiple tasks with a single model \citep{demsar.2006}, rather than each task individually. We leave consideration of  this as an interesting direction for future work.}

\section{Machine Translation} \label{sec:mt}

To show how our approach to power analysis can be applied to
a more difficult setting, we consider
automated
evaluation of machine translation using BLEU
scores
\citep{papineni.2002}.
As
with 
accuracy, we would like to know 
what scale of improvements can be detected with reasonable power on typical test sets.
This setting is more complicated 
because
(1) BLEU is a corpus-level metric, rather than being averaged across instances, 
and 
(2) typical models are trained on vast amounts of parallel data, with little
data
available that has not been used in training, making it difficult to
estimate variation in performance. 

\paragraph{Significance testing for BLEU:} To test for a significant difference between two MT models
we use the
randomization test, as recommended in \citet{dror.2018}: given the paired output translations from both models, 
swap the outputs
for a random subset of test examples and compute the resulting difference in BLEU.
Repeating this thousands of times 
gives us a null distribution, which can be used to test the observed difference between models.

\paragraph{Generative process for simulations:}
If large amounts of untouched evaluation data were available, we could approach power analysis by simply evaluating BLEU score on many random subsets of $n$ sentences, and computing the mean and variance of each system. Unfortunately, because MT depends on parallel text (most of which is used in training), evaluation data tends to be scarce. Instead, we introduce a generative process that can produce the necessary inputs for power analysis.

For intuition, note that if we swap the $i^{\textrm{th}}$ pair of model outputs (as is done in the randomization test), leaving rest as they are, we change the difference in BLEU between models by a specific amount, $\delta_i$, which we call the effect of making that swap.
While these individual effects are not independent of each other due to the corpus-level nature of the metric, in practice, the sum of individual effects closely approximates the net effect of  swapping entire subsets (see Figure \ref{fig:mt_all_correlations} in   Appendix~\ref{app:mt}).

Based on analyzing several models and datasets, we find the typical distribution of these individual effects can be approximated using a mixture of a Delta distribution at zero, and a Laplace distribution (see Appendix~\ref{app:mt} for details).
Concretely, 
if we assume $\Delta_B$ is the expected difference in BLEU between two models on a dataset of $n$ examples, 
and $P_0$ is the expected proportion of examples for which $\delta_i=0$, 
we can simulate a dataset  $\{\delta_i\}_{i=1}^n$ of $n$ individual effects using the following process:
with probability $P_0$, $\delta_i = 0$. With probability $1 - P_0$, $\delta_i \sim \textrm{Laplace}(\mu, b)$, where $\mu=\frac{-2  \cdot \Delta_B}{n  (1-P_0)}$,  $b=b_0/n$,
and $b_0$ is a user-specified parameter that controls the variance, independent of the sample size.
By construction, $\mathbb{E}[\sum_{i=1}^n \delta_i] =  -2 \cdot \Delta_B$.\footnote{Note that swapping all $n$ examples would reverse the model scores, equivalent to a net effect of $-2 \cdot \Delta_B$.}

Given this generative process, we can then estimate power using the Algorithm in Figure 2. On each iteration, draw a simulated dataset from the generative process, compute the observed difference between models as $\hat \Delta_B = -\frac{1}{2} \sum_{i=1}^n \delta_i$, and test if this is significantly different from zero using a modified randomization test, in which we assume that the net effect of swapping a subset of instances is simply the sum of the $\delta_i$'s in the subset. (Please see online materials for an interactive example). 

\paragraph{Empirical estimates:}

\begin{table}[]
    \centering
    \small
    \begin{tabular}{l l c c c c c c}
        \toprule
        \textbf{M1} & \textbf{M2} & \textbf{Test set} & $n$ & $\Delta_{\textrm{B}}$ & $\hat P_0$ & $\hat b_0$ \\
        \midrule
        TF19$^*$ & TF18$^*$ & 2019 & 2K & 4.3 & 0.19 & 23.7 \\
        TF18 & TF16 & 2018 & 3K & 4.2 & 0.09 & 29.4 \\
        TF16 & Conv17 & 2017 & 3K & 1.3 & 0.12 & 22.5 \\
        TF16 & Conv14 & 2016 & 3K & 7.6 & 0.10 & 27.6\\
        \bottomrule
    \end{tabular}
    \caption{Relevant parameters from four MT evaluations. TF are Transformer-based 
    \citep{ott.2018,edunov.2018,ng.2019} and Conv are Convolutional models
    \citep{gehring2017convolutional} from \textsc{fairseq}. Test sets are from WMT shared tasks for En-De translation. $\Delta_B$ is the reported difference in BLEU, whereas $\hat P_0$ and $\hat b_0$ are estimated.  * indicates ensembles.}
    \label{tab:mt_params}
\end{table}

In order to estimate reasonable values for the required parameters, we use several pretrained models from the \textsc{fairseq} library \citep{ott.2019} for the WMT English-German translation task.
We evaluate these models on the shared task test sets from 2016-2019 and compute BLEU scores using \textsc{sacrebleu} \citep{post.2018}. 
Fitting a Delta-Laplace mixture to the effects of swapping individual output pairs, we estimate values for $\hat P_0$ and $\hat b_0$, reported in Table \ref{tab:mt_params}. (See also Figure \ref{fig:mt_all_laplace} in Appendix \ref{app:mt}; code for computing estimates is provided in the online materials).

While far from identical, the four comparisons, each representing different stages of model evolution, all produce similar estimates.
Although these estimates are only based on a single language pair, the models and test sets are relatively diverse, and we expect that these estimates will generalize, though better estimates could be obtained by fitting this distribution to a new domain of interest.

\begin{figure}[]
    \centering
    \includegraphics[width=0.99\linewidth]{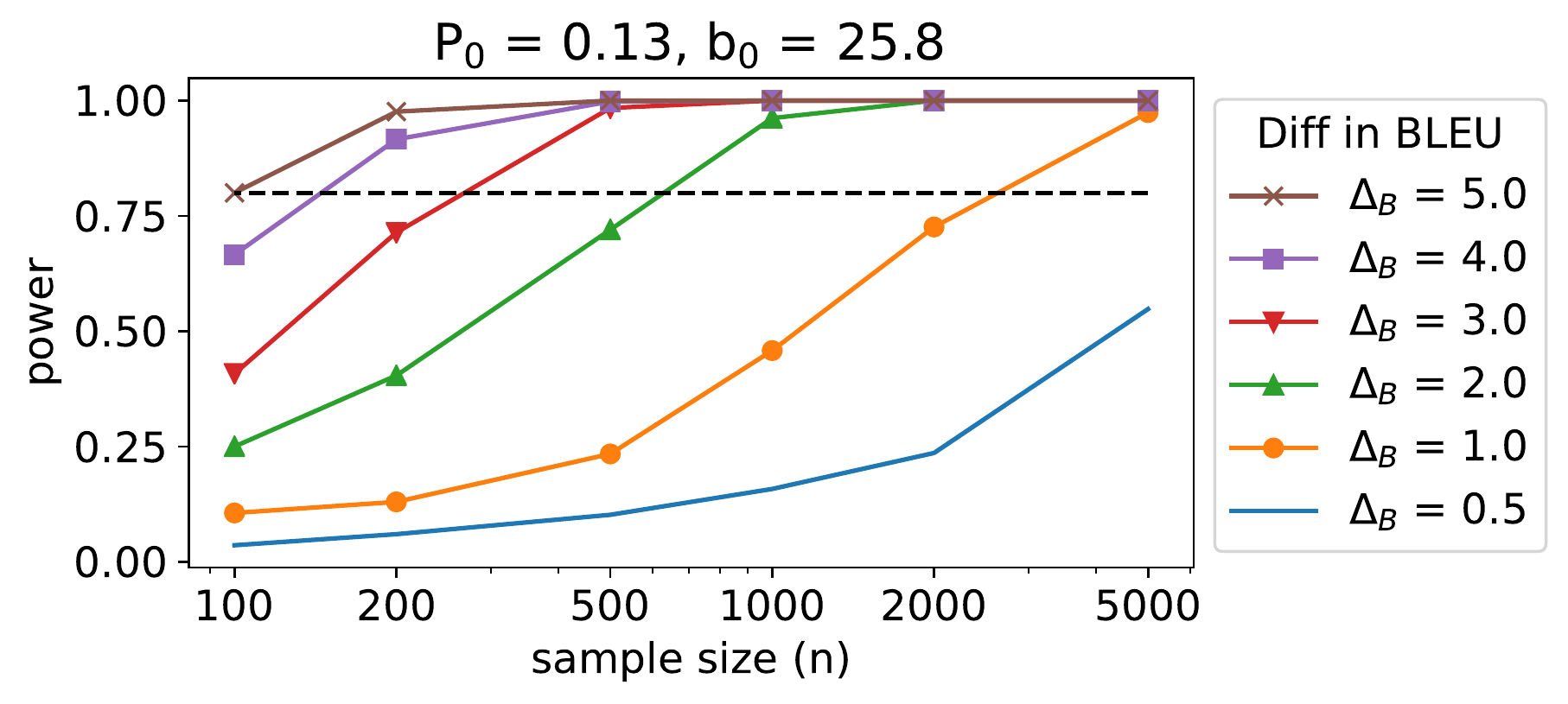}
    \caption{Power analysis for MT, 
    showing how
    power increases
    with $n$ and $\Delta_B$, 
    using 
    an average of fitted values for $P_0$ and $b_0$.
    Based on this analysis, we expect that an experiment with a test set of 2000 sentences would have approximately 75\% power to detect a difference of 1 BLEU point 
    as significant. For additional plots, refer to Figure \ref{fig:ref_mt_p0_and_scale} in Appendix \ref{app:mt}.
    }
    \label{fig:mt_power_dbleus}
\end{figure}

Using these estimates, we can now
characterize how much power test sets of different test set sizes ($n$) would have for a range of possible differences in BLEU ($\Delta_B$).
Figure~\ref{fig:mt_power_dbleus} shows this for $P_0$ and $b_0$ set to the average of the observed values.\footnote{For a sensitivity analysis of how power varies under different assumptions for $P_0$ and $b_0$, please see Figure \ref{fig:ref_mt_p0_and_scale}
in Appendix \ref{app:mt}.}
Based on this estimate, we conclude that
for typical MT test sets of around 2,000 examples,
an improvement of 1 BLEU point can likely be detected with approximately 75\% power.
As shown in 
Figure~\ref{fig:mt_power_dbleus} 
this power level increases dramatically with sample size and effect size.

This analysis has served, in part, to show how a simulation-based approach to power analysis can be adapted to virtually any task. Additional work is required to test how well these specific parameter estimates will generalize, but the same process can easily be adapted to new language pairs. More generally, there would be great value in the MT community curating larger held-out test sets, both to validate this analysis, and for better powered future comparison.

\section{Likert-Scale Human Evaluations}
\label{sec:humaneval}

Tasks such as natural language generation are difficult to evaluate using 
automated methods; as such, human evaluations 
are
central to
NLP. 
Past work has reported
great variation in how
human evaluations are done
\citep{van2019best}. Therefore, we 
begin with
a meta-analysis of a subset of human evaluation experiments from EMNLP 2019, which we then use as the basis for claims about the power of human evaluations in NLP more generally.

\subsection{Meta-analysis}
\label{sec:meta}

To characterize the state of human evaluation in NLP, we identified papers from the main session of EMNLP 2019 that made use of human evaluations (details in Appendix \ref{app:humandata}). To generalize across studies, we restrict 
our analysis
to 
Likert-scale comparisons, which was the most commonly reported type of evaluation. We extracted all 
cases where a new model was being 
compared to
the best-performing baseline on one more metrics
(117 comparisons from 41 papers)
and  normalized all ratings to be on a 0-1 scale.

One
takeaway from this meta-analysis 
is that the
reported effect sizes
(that is, difference between the novel model and the best-performing baseline)
vary widely ($\textrm{s.d.}=.12$ on a [0, 1] scale).
Number of items tested is more consistent: 69\% used 100 or fewer, and only 18\% used over 200. But, as similarly found by \citet{van2019best}, many key details were not reported in this sample of experiments.
Most commonly missing was number of ratings per item (34\% of all experiments),
followed by total number of workers
(28\%). 
For 7\% of experiments, we could not determine the number of items tested. 
57\% of experiments collected 3 annotations per item, which was also the modal number of unique annotators. Thus, it is often difficult to ascertain, for any particular experiment, the details of the experimental setting that are necessary to evaluate the validity of the results.

\begin{figure}[]
    \centering
    \includegraphics[scale=0.62]{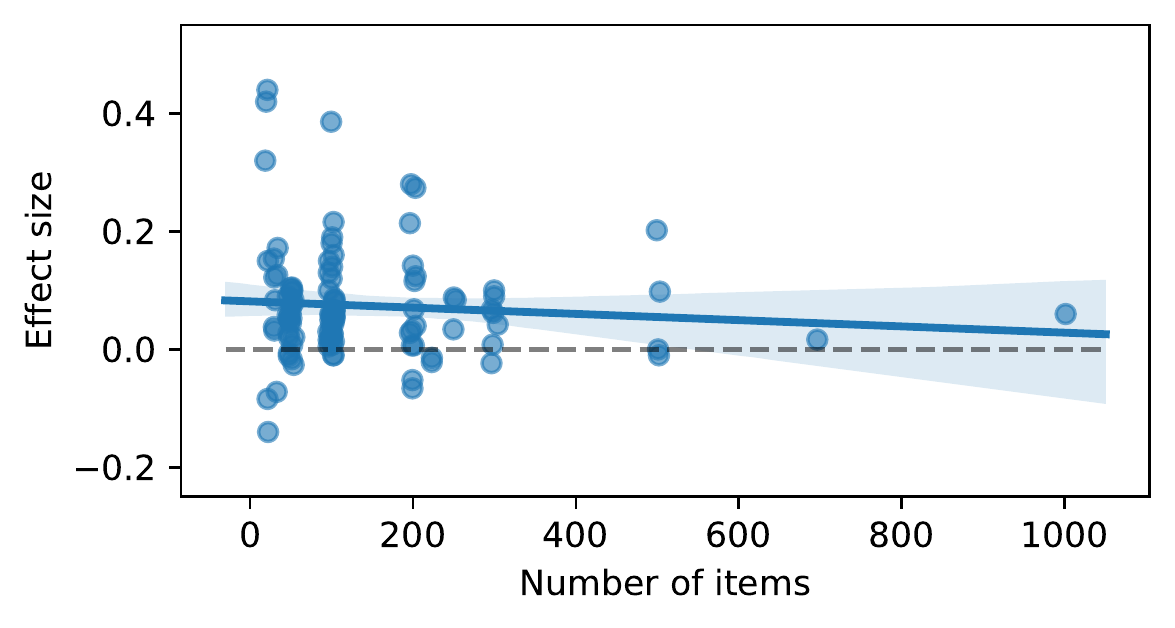}
    \caption{Scaled effect size vs.~number of items from our EMNLP 2019 survey,
    showing
    higher variance in the smallest
    samples.
    There is a slight negative correlation, though it is not significant.
    As can be seen, most experiments are small ($n \leq 100$).
    }
    \label{fig:emnlp}
\end{figure}

Because the number of items rated was the most commonly reported, we use that as our proxy for sample size. Figure \ref{fig:emnlp} shows
scaled
mean
difference between models
as a function of
number of items. 
As expected, we see greater variance in effects with smaller samples since, with smaller samples, we expect greater noise. We also observe a slight negative correlation between effect size and sample size. That is, as sample size gets larger (and, thus, as estimates get more precise), the estimated effect size gets smaller. This trend is sometimes used as an indication of publication bias (censoring of null and opposite-direction effects) since, in a sample with no publication bias, the effect size should be independent of the sample size \cite{begg1994operating}. However, in our case, this correlation is not significant (Kendall's $\tau = -.07$, $p = .32$) and so it is difficult to draw strong conclusions.\footnote{We exclude from this analysis two large negative effects with $N=500$ which would exaggerate this correlation.}

\subsection{Power analysis for human Likert ratings}
 
What kind of effect sizes can  typical human evaluation experimental designs detect? 
As in previous sections, we can use simulations to explore how many
annotators and/or instances
should be
used
to have sufficient power.

Simulating human experiments is conceptually simple (e.g., $m$ raters each rate $n$ generated sentence on overall quality), but for realistic simulations, we need to consider variation in items (some generated sentences are better than others), and variation by rater (some raters use higher ratings and/or respond to different aspects of quality), as well as the overall difference in quality between models. A simulation which treated all workers as identical would fail to capture this variation, and hence might overestimate power \cite{barr2013random}.

Unfortunately, details such as worker
variance
are
rarely reported in published papers.
To better characterize the typical variation in human evaluations, 
we rely on a convenience sample of several large datasets to estimate these parameters and use them in our simulations as a proxy for what we might observe in practice.
Although 
focused on different tasks, 
all use a similar methodology, namely, getting many Likert-scale annotations per instance from 
many 
annotators and models (in some cases as many as 20 ratings per item).\footnote{We use publicly available or author-provided data from  \citet{hashimoto2019unifying,dathathri.2020,holtzman.2020}, and WMT19 (links in Appendix \ref{app:humandata}).}

In order to extract estimates of these parameters for our simulations, we use  hierarchical mixed-effects models, as used in psychology and other behavioral fields \citep{barr2013random,gelman2006data}.
Such models incorporate 
variation in 
the quality of generated instances, 
annotator responses, 
and annotator sensitivity, and are recommended by \citet{van2019best} for analyzing human evaluations. (We provide details in Appendix \ref{app:lmer} and include code for fitting such models as part of the online materials).
Using this approach, we obtain an estimate of the relevant parameters from each of the large datasets.
From these, we choose sets of 
parameters 
to be representative of experiments with high or low variance, with full results in Appendix \ref{app:lmer} (see Table \ref{table:lmer_recs} for parameter estimates).

\begin{figure}[]
    \centering
    \includegraphics[scale=0.52]{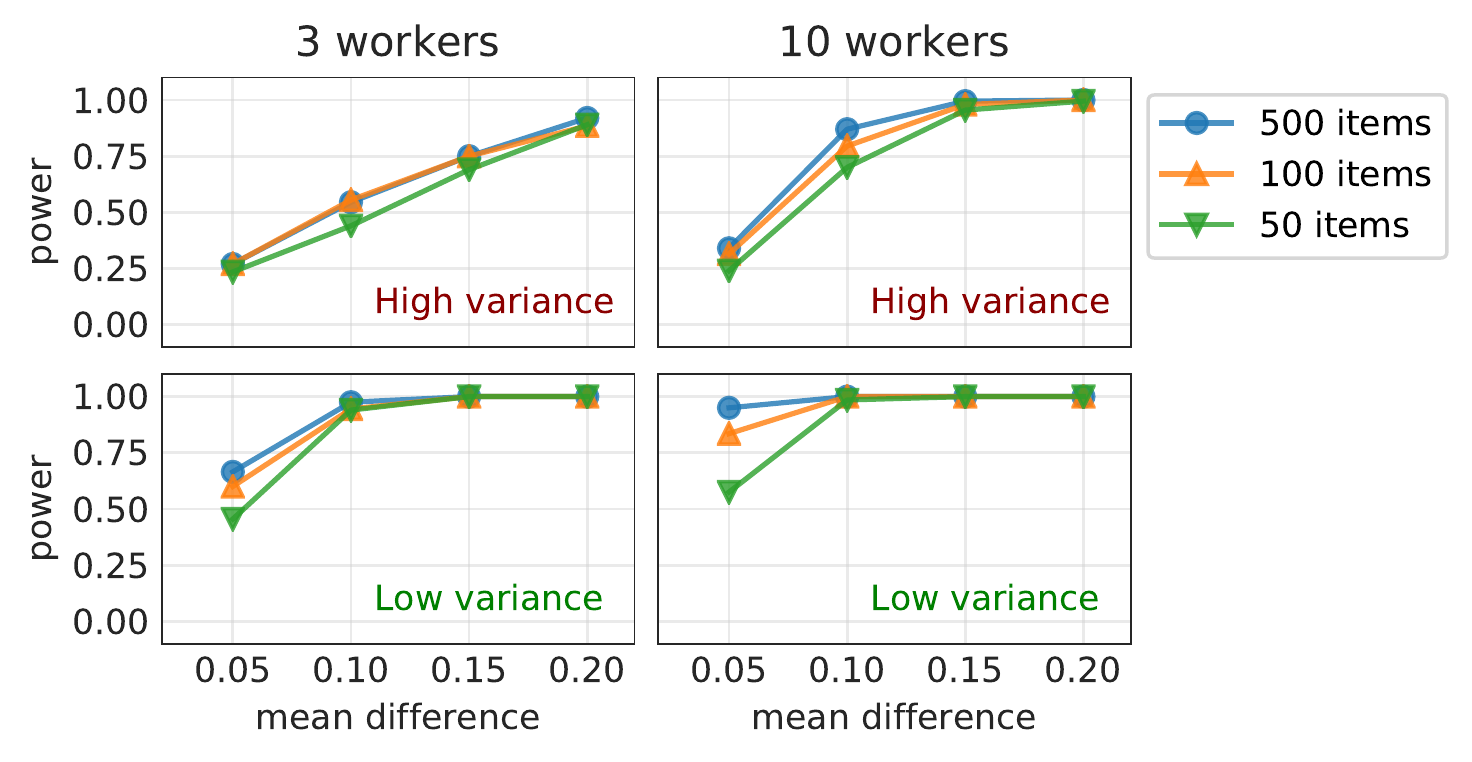}
    \caption{Using parameters estimated with mixed effects models from a high variance setting (top) and a low variance setting (bottom), the left panel shows simulated experiments with 3 workers annotating each item, the right panel shows an unusually high number of annotators per item (10 workers). Under typical assumptions, many common experimental settings (e.g., 3 workers and 100 items) are underpowered.}
\label{fig:conservativelmer}
\end{figure}

As before, we then use these estimates to simulate data, assess significance on the simulated data (here using mixed effect regression), and compute power as a function of mean difference and sample size.\footnote{These simulations require estimates for 7 parameters: the baseline, the effect size, variance by worker, variance by worker as a function of model, variance by item, variance by item as a function of model, and residual variance.} The resulting power estimates are shown in Figure \ref{fig:conservativelmer}, plotted in terms of effect size, sample size, and numbers of workers and items, for both the high and low variance scenarios. 
From this analysis, we highlight a few key takeaways:
\begin{itemize}
\item \textit{
Many
human evaluation studies are likely underpowered:} 
Using the ``high variance'' parameters (which are typical of most of the datasets we used), the most common design at EMNLP 2019 (3 workers, 100 items)
is underpowered
unless the effect size is quite large (0.2 or higher on the [0, 1] scale).
\item \textit{Even with low variance, typical designs are underpowered to detect small effects:}
Using our estimated parameters for the low variance setting, experiments will be underpowered to detect small effects (0.05 on the [0, 1] scale), unless an unusually large number of ratings per item are collected (10+ for 100 items).
\item \textit{Need for improved reporting:}
Most human evaluations do not report enough detail to interpret the results. This could be drastically improved through basic power analyses, significance testing using mixed-effects models, and sharing of raw data.
\end{itemize}

Given our model estimates and simulations, we conclude that, in aggregate, many human evaluations are underpowered and would benefit from larger sample sizes, particularly by using more workers per item. Increased adoption of even approximate power calculations within the NLP community will promote thoughtful consideration of appropriate sample sizes and improve the reliability and replicability of results.

\section{Overall Recommendations}

\begin{itemize}
    \item
    Power analyses should be done prior to evaluation when comparing against a baseline.
    If a comparison is likely to 
    be underpowered, the pros and cons of running that evaluation should be carefully considered. Underpowered experiments do not provide convincing evidence of progress.
    \item For new datasets and shared tasks, the number of instances in the test will determine the minimum detectable effect size, and should be chosen accordingly.
    \item For 
    tasks which no longer have adequate power to detect typical improvements
    (e.g., MRPC and SST-2), authors should consider expanding the test set or retiring the task.
    \item To facilitate future power calculation and significance tests, model owners should release final fine-tuned model checkpoints.
    Alternatively, leaderboard owners may wish to make validation set predictions from all submitted models publicly available. 
    \item For human evaluations,
    (anonymized) raw data should be shared, along with parameters and code to replicate the analysis,  including proper significance testing.
    Prior to collecting human evaluation
    data, researchers should create an analysis plan and run power analyses to determine an appropriate sample size (likely requiring more workers and items than is currently typical in NLP).
\end{itemize}

\section{Conclusion}

Recent progress in NLP has been extraordinarily rapid, sometimes at the cost of experimental rigor. In this paper, we have presented evidence that underpowered experiments are widespread in NLP.
For comparisons based on small samples, there is little reason to think that such an evaluation \textit{could} reliably provide evidence of a significant improvement, and good reason to believe that improvements found to be significant will exaggerate or reverse the true effect.  
Going forward, a combination of larger test sets, simple power analyses, and wider sharing of code, data, and experimental details will help to build the foundation for a higher standard of experimental methodology in NLP.

\section*{Acknowledgments}

Toyota Research Institute (``TRI'')  provided funds to assist the authors with their research but this article solely reflects the opinions and conclusions of its authors and not TRI or any other Toyota entity. Thanks to Sam Bowman, Amanpreet Singh, Kevin Clark, Naman Goyal, and Colin Raffel for providing data from submissions to the GLUE leaderboard, as well as Taylor Berg-Kirkpatrick, Sumanth Dathathri, Ari Holtzman, Hannah Rashkin, and Nikita Srivatsan for providing raw human evaluation data, not all of which made it into the paper.

\bibliography{refs}
\bibliographystyle{acl_natbib}

\newpage
\appendix

\section{Further Discussion of Significance Testing, Power Analysis, and Post-Hoc Analysis}
\label{app:nhst}

\paragraph{Null hypothesis significance testing:}
In this paper, we work within the framework of null hypothesis significance testing (NHST). NHST is not free from problems, in that certain systematic processes within the practice of scientific research and publishing can undermine its advantages, many of which have been explored in the literature \citep{gelman.2013,ioannidis.2019,mcshane.2019}.
Nevertheless, it would be premature to discard the entire paradigm, and we believe there is still some value in considering power within NHST for several reasons.

First, despite its flaws, NHST remains a commonly used experimental framework in NLP research. 
Whether implicit of explicit, most experimental comparisons in the NLP literature have the structure of an experiment in the NHST framework, where having equivalent performance to an existing baseline is treated as a null hypothesis and the new model is argued to be significantly better (the typical case) or significantly worse (far rarer). But, whereas many fields that run experiments have standardized procedures for assessing statistical significance, NLP papers vary as to how formally they use a hypothesis testing framework to evaluate their results \cite{berg.2012,van2019best,azer.2020}. 

Second, when done properly, NHST does provide a convenient way of summarizing results.
Improvements in overall methdology, such as sharing code and data, sensitivity analyses, greater interest in null findings, and even pre-registration can vastly improve the validity of this paradigm, and we are seeing adoption of some of these practices within NLP. 

Finally, there is also a great need for additional clarity with respect to precisely what claims are being made by NLP papers. 
In this work, we are primarily focused on claims made about trained models (i.e. in testing whether one particular instantiation of a model is significantly better than a particular instantiation of another model). It is, of course, also important to consider broader claims that might be made, such as about expected performance or computational budget \citep{dodge.2019,schwartz.2019}, and everything we have to say can be extended to incorporate such considerations. For the purpose of clarity, however, we restrict ourselves to the simplest sort of statistical claim.

\paragraph{Power and power analyses:}
The probability that a statistical test will reject the null hypothesis in an experiment is a function of several parameters, some of which are typically known or controllable, such as the sample size and significance threshold, and some of which are unknown, such as the details about exactly how models differ. Power tells us what this probability would be, if we knew the true values for these unknown parameters. 
Conditional on a particular difference existing (e.g. an expected difference in accuracy between two models for a particular data distribution), along with a statistical test, a significance threshold, power is the probability that the test will reject the null hypothesis and find the observed difference to be significant. In common statistical terminology, power is one minus the probability of false negatives in rejecting the null hypothesis or type II error.

While we will not, in general, know what the true power of an experiment is, by making reasonable assumptions, we can try to choose appropriate values for those parameters that we can control. By making assumptions about what we expect to observe, we can obtain estimates of how much power a test is likely to have, which may lead us to modify our experimental design, such as by increasing the sample size.

Importantly, proper experiment design requires specifying these parameters in advance of data collection, or otherwise using a valid stopping rule.  One can \emph{always} obtain a significant result by progressively collecting data until a significant result is found (``sampling to a foregone conclusion''), but this is not a valid procedure
\citep{anscombe.1954,wagenmakers.2007}. Similarly, \emph{post-hoc} power analysis, using estimates derived from the experiment itself, provides no additional information beyond a transformation of the observed $p$-value, and is thus not recommended (though see below).

Expanding on the algorithm in Figure \ref{fig:algo}, a simulation-based power analysis involves the following:
\begin{enumerate}
\item First, determine the statistical test, $T$, which will be used. For the example of comparing models depicted in Figure \ref{fig:ABdemo}, we will use the binomial test to compare the systems \citep{dror.2018}.
\item Come up with a generative process which could be used to generate data like that which we will collect. In this step, we need to make assumptions about the comparison of interest. Since the binomial test requires only the counts of how many people prefer each system, we need to specify a prior on generating those counts. For example, we might assume that 60\% of people will prefer system B, so the generative process will be $c_B \sim \textrm{Binomial}(p=0.6, n)$, where $n$ is the total number of people to be sampled.
\item Choose a value of $n$ for which we want to calculate power. Repeatedly (e.g., 10,000 times) draw many samples from our assumed generative process for that size of $n$ . 
\item For each simulated dataset of size $n$, run the chosen statistical test to check if difference between the observed counts is significant, and compute the proportion that are found to be significant. This is our estimate of power.
\end{enumerate}

Note that more direct solutions for power analysis do exist for some settings, such as this one (see Appendix \ref{app:binomial} below). 

\paragraph{Post-Hoc Power Analysis:}
Post-hoc power analysis is an issue when the true population effect has variance to it ~\citep{okeefeposthoc,theabuseofpower,gelman_dont_2019}. In the case of NLP models, there are several perspectives on the comparisons which can lead to differences regarding how we perceive post-hoc power analysis: (1) we are comparing one model vs. another on a particular test set, the effect we see is the true population effect, post-hoc power analysis is okay because it is deterministic; (2) we are comparing one model vs. another on a data distribution from which the test and dev set are drawn, post-hoc power is not okay; (3) we are comparing one training algorithm vs. another (including variance from both training procedures and test/dev set draws), post-hoc power analysis is still not okay. We specifically look at the case of (2). While (3) is interesting on its own, this is not the typical comparison done (yet) in NLP research and thus we do not have enough information on reported training variance to investigate this thoroughly here. The case of (1) is also atypical as the authors of a study typically wish to draw inferences about how well a model does on the true data distribution (hence, why a dev and test set are used).

\section{Type-M and Type-S errors}
\label{app:typem}

Although the most obvious risk of using underpowered experiments is that there is a greater chance of failing to detect a true effect, there is an additional harm of using an underpowered design, which has emerged in light of the replication crisis in science. This can be most easily understood through the idea of Type-M and  Type-S error \citep{gelman2014beyond}.

Type-M error is the extent to which an observed difference exaggerates the true effect, conditional on a finding being significant. Type-S error is the probability that an observed difference has the opposite sign of the true difference, again conditional on a finding being significant. Even in a low-powered experiment, there is some probability of finding an effect to be significant; the lower the power, however, the more likely it is that the observed significant difference has the opposite sign of the true effect, and the larger the degree to which the magnitude of the observed effect will tend to exaggerate the true effect.

Intuitively, if power is low, this means that the sample size is small relative to the effect size. As such, the difference will \emph{only} be significant if an atypically large effect is observed. Assuming the use of a two-sided test, many of these significant findings will also have the wrong sign, as they will be nearly as likely to fall on either side of zero for a symmetric distribution.

Type-M and Type-S error rates can be estimated using the exact same process for power analysis as described in Figure \ref{fig:algo}. To do so, we need only augment the algorithm with these two additional steps:
\begin{enumerate}
\item[3.] $\textrm{Type-S error} \approx \sum_{i : p_i \leq \alpha} \frac{\mathbb{I}[\textrm{sign}(e_i) \ne \textrm{sign}(e*)]}{| {j : p_j \leq \alpha} |}$
\item[4.] $\textrm{Type-M error} \approx \sum_{i : p_i \leq \alpha} \frac{\textrm{abs}(e_i) / \textrm{abs}(e^*)}{| {j : p_j \leq \alpha} |}$
\end{enumerate}

\begin{figure}[!ht]
    \centering
    \includegraphics[scale=0.53]{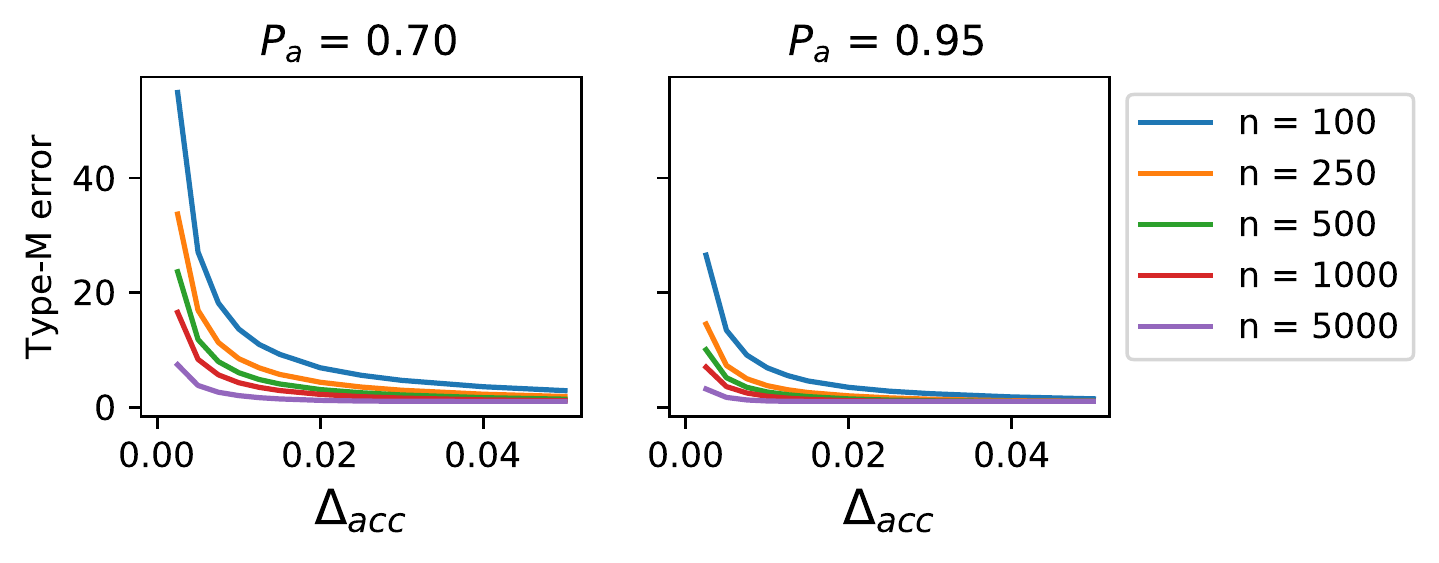}
    \caption{Type-M error (the factor by which observed significant effects are likely to exaggerate the true effect) for comparing classifiers on accuracy increases with smaller test sets ($n$), smaller differences between models ($\Delta_{acc}$), and smaller agreement rates ($P_a$). Severe exaggerations of differences between models  are likely with underpowered designs. }
    \label{fig:typem}
\end{figure}

\begin{figure}[!ht]
    \centering
    \includegraphics[scale=0.53]{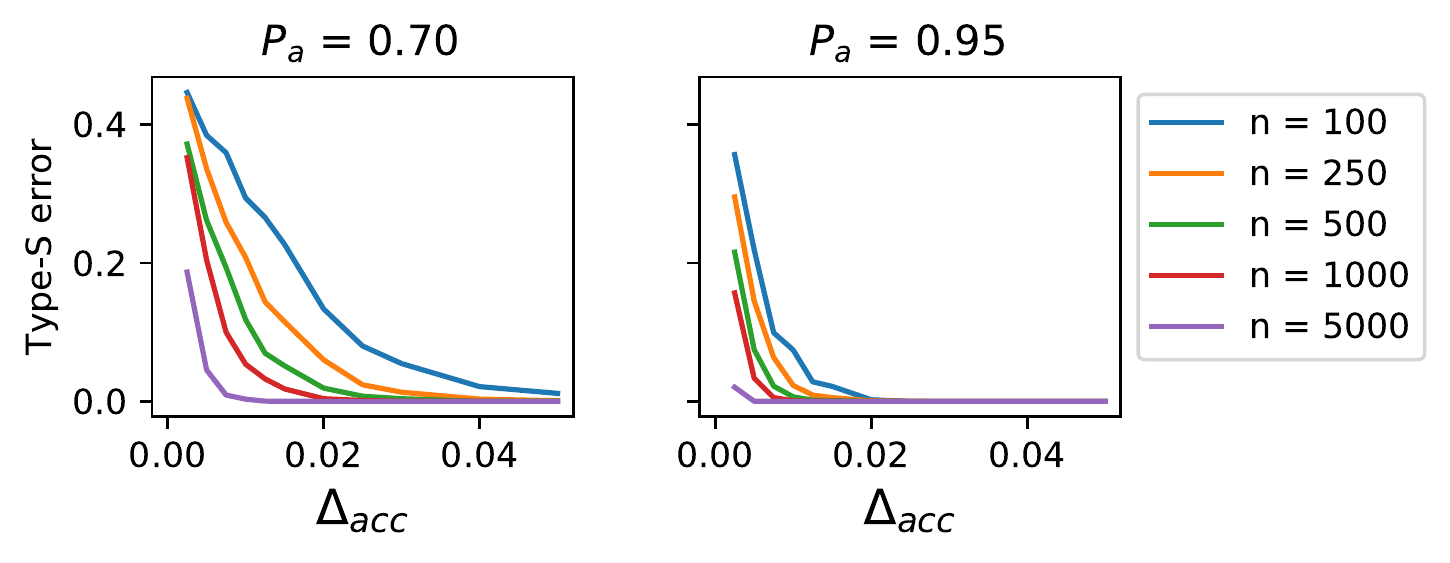}
    \caption{Type-S error (the probability that significant differences observed between models will have the opposite sign of the true difference) for comparing classifiers increases with smaller test sets ($n$), smaller differences between models ($\Delta_{acc}$), and smaller agreement rates ($P_a$). Sign errors become resaonably likely with underpowered experiments.}
    \label{fig:types}
\end{figure}

Figures \ref{fig:typem} and \ref{fig:types} show scenarios for comparing classifiers on accuracy, corresponding to Figure \ref{fig:ref_acc} in the main text, but showing expected  Type-M and Type-S error instead of power. As can be seen, Type-M and Type-S error increase with smaller sample sizes, smaller differences between models, and lower agreement rates, all corresponding to lower power.

\section{Numerical Example of a McNemar's Test Simulation}
\label{app:mcnemar_example}
To provide a concrete example of comparing classifiers on accuracy, imagine that a test set for a benchmark task has 500 instances. Based on prior knowledge (see main paper), we might assume that our proposed model will achieve, at most, an absolute improvement of 2 percentage points over the state of the art ($\Delta_{acc}=0.02$), and that the models are likely to agree on 90\% of examples ($P_a=0.9$). We can convert these assumptions into a distribution over outcomes which will define our generative process. In particular, for a random unseen instance, these assumptions imply that there is a 10\% chance of a disagreement; the probability that our model is correct and the old model is incorrect is therefore 6\%, and the opposite outcome has a probability of 4\% (giving us the assumed net difference of 2\%). Note that, because McNemar's test does not consider the on-diagonal elements, it is not necessary that we explicitly define the baseline accuracy. Thus, a valid probability distribution for use in this simulations could be that shown in Table \ref{tab:contingency_example}.

\begin{table}[]
\begin{center}
\begin{tabular}{l|cc}
        & M1 correct & M1 incorrect\\\hline
M2 correct & $0.6$ & $0.06$\\ 
M2 incorrect & $0.04$ & $0.3$\\ 
\end{tabular}
\end{center}
    \caption{A possible distribution corresponding to the case where models M1 and M2 will agree on 90\% of examples ($P_a$) and M2 achieves a 2\% improvement over M1 ($\Delta_{acc}$). Note that the on-diagonal terms here will be dictated by the accuracy of M1 (or equivalently, by M2), but for our purposes, only need to be non-negative and sum to $P_a$ for the sake of McNemar's test, which only looks at the off-diagonal elements.}
    \label{tab:contingency_example}
\end{table}

By drawing many samples from this distribution of size $n=500$ and computing a $p$-value using McNemar's test for each, we obtain an estimate that the power of this test is approximately $0.25$ for a significance threshold of $\alpha=0.05$, which is severely underpowered. This would also imply a Type-M error factor of 1.9; we would expect that a typical experiment that found the observed difference between models to be significant would exaggerate the true difference of 0.02 by a factor of 1.9, producing observed significant differences between models on the order of 0.04, on average. (See supplementary notebooks for calculations and  interactive demonstration). As such, we conclude that this test set is too small to be able to  reliably evaluate whether or not our model is significantly different from the state of the art, and should distrust any observed differences that are significant, unless we have poorly estimated the relevant parameters.

By contrast, if the test set contained 2000 examples, we would estimate the test to have nearly 80\% power, with a Type-M factor of only 1.1, and would feel comfortable proceeding with and reporting on this evaluation. Similarly, if we had reason to think that our model represented a game-changing advance, and would achieve an improvement of 4 percentage points, or if we had reason to believe that the models would agree on 97.5\% of examples, then we would have the power to evaluate this, even with only 500 examples.

\section{SQuAD 2.0 Analysis and Results}
\label{app:squad}

From the authors of SQuAD 2.0, we obtained pairwise agreement statistics on the SQuAD 2.0 development and test sets for all models that were submitted to the SQuAD 2.0 leaderboard 
and had publicly visible development set predictions on the CodaLab platform. 
We removed six submissions whose exact match (EM) scores on test data were less than $50\%$; 
EM scores below $50\%$ suggest a bug or misconfiguration of the model for predicting on the test set, 
as the majority baseline gets roughly $50\%$ accuracy (by always predicting no-answer).
We also removed one submission whose development set EM score was more than $20$ points higher than its test EM score, as it seemed likely that the model had been trained on the development set.
After this filtering, we were left with 144 models.

\begin{figure}[ht]
    \centering
    \includegraphics[scale=0.6]{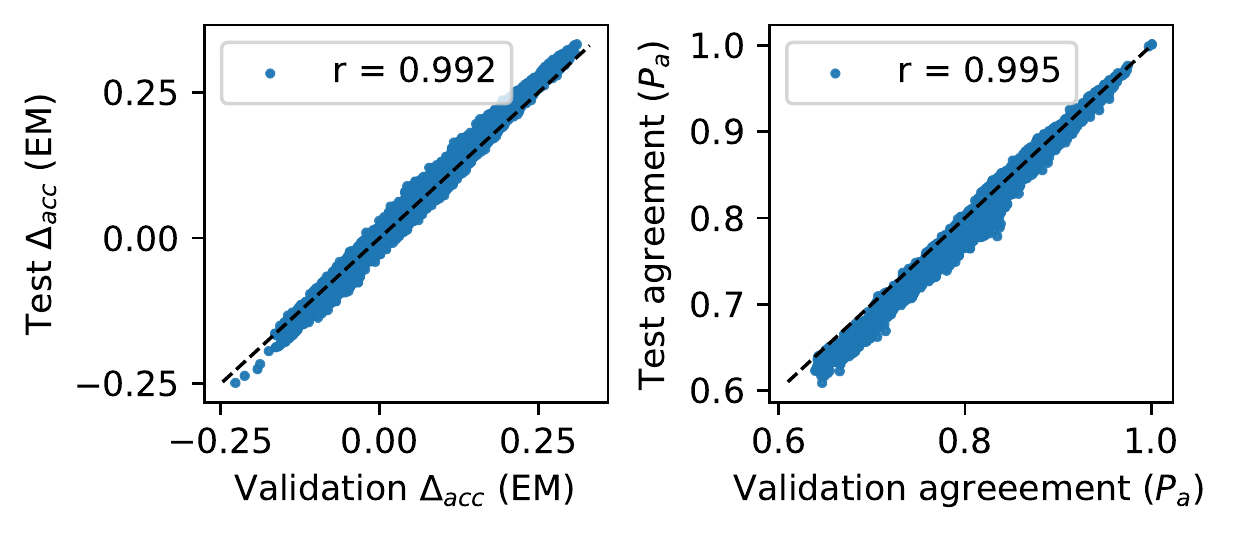}
    \caption{Correlation between validation and test data among all models submitted to the SQuAD 2.0 leaderboard for both pairwise accuracy differences ($\Delta_{acc}$ using exact match (EM); left), and agreement rates ($P_a$; right). In both cases, Pearson correlation ($r$) is over 0.99. Dashed lines show $y=x$.}
    \label{fig:squad_correlation}
\end{figure}

Figure \ref{fig:squad_correlation} shows the correlation between validation and test data for both pairwise accuracy differences ($\Delta_{acc}$) and agreement rates ($P_a$) on the SQuAD 2.0 leaderboard. As can be seen, these correlate well, suggesting that measuring these quantities on validation data can serve as a reasonable guide when doing a power analysis for a new model, though lower agreement rates on dev data to tend to slightly underestimate agreement on test. If the validation results are available for both models, these can be used to compute estimates of $P_a$ and $\Delta_{acc}$, and these can be used to compute the approximate power of the test set.

\begin{figure}[ht]
    \centering
    \includegraphics[scale=0.47]{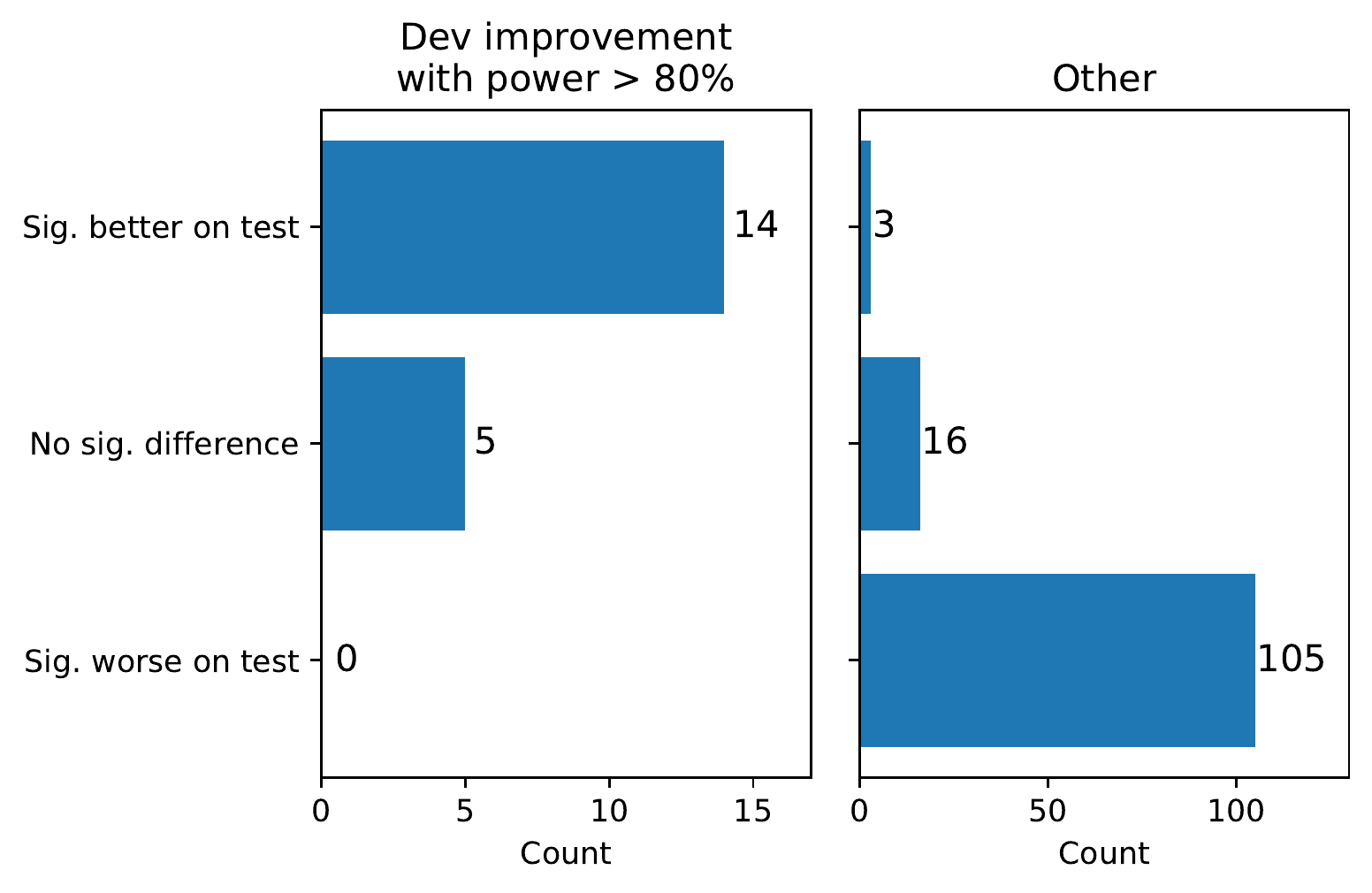}
    \caption{SQuAD 2.0 leaderboard submissions compared to previous SOTA, where we require for SOTA that submissions have 80\% power (based on validation improvement and agreement), and a significant improvement on test data.}
    \label{fig:squad_control}
\end{figure}

To verify that using these estimates provide a reliable guide to power, we make use the predictions made by SQuAD 2.0 submissions on both validation and test data. 
In particular, if we assume that each submission is being compared to the previous model to demonstrate a significant and well-powered improvement over the previous baseline, we find that 19 out of 143 submissions showed sufficient improvement on the validation set to have at least 80\% power (see Figure \ref{fig:squad_control}). Of these, 14 (74\%) attain a significant improvement over the baseline on the test data (consistent with the expected value of 80\%). Of the remaining 124 submissions, 3 (2.5\%) would show a significant improvement over the baseline, but did not have sufficient power based on validation performance. Interestingly, while all other significant improvements were generally well-spaced over time, these three underpowered submissions were all beaten by a new submissions within 5 days. As an aside, we also note that the vast majority of submissions are significantly worse than the current SOTA, reinforcing the notion that real improvements are rare, and most improvements will be small.

\paragraph{Caveats:} Correlation between the effect size on the validation and test sets may not always be so high. Overconfidence in the power of your experiment may thus occur if the validation performance is greater than the test performance (as would be the case if no regularization was used and extensive hyperparameter tuning caused a model to overfit to the validation set). Alternatively, if comparing to a baseline with inflated performance on validation data (for the same reasons as above), running power analyses based purely on estimates from validation data would underestimate power. As such, combining validation estimates with reasonable priors is recommended.

\section{Accuracy}
\label{app:mde_table}

\subsection{Data Collection}

\subsubsection{Model Predictions on Test Set and Model Prediction Agreement}
From the authors of the GLUE benchmark -- as well as authors of individual models -- we obtain the model test-set predictions on all tasks from a set of 10 high-performing models, which allows us to measure the extent to which their predictions overlap with each other. We select GLUE tasks which use accuracy as an evaluation metric. The relevant tasks are MNLI~\citep{williams2018broad}, MRPC~\citep{dolan-brockett-2005-automatically}, RTE~\citep{dagan2005pascal,bar2006second,giampiccolo2007third,bentivogli2009fifth}, SST-2~\citep{socher2013recursive}, QQP~\citep{iyer2017first}, QNLI~\citep{rajpurkar2016squad}, and WNLI~\citep{levesque2012winograd}. For consideration of other metrics, see Appendix \ref{app:metrics}. 

We use model predictions for: ELECTRA (small, base, large, large with tricks)~\citep{clark2019electra}, XLNet (large)~\citep{yang2019xlnet}, T5~\citep{raffel2019exploring}, ALBERT (large)~\citep{lan2019albert}, BAM(large)~\citep{clark2019bam}, RoBERTa (large)~\citep{liu2019roberta}, and BERT~\citep{devlin2018bert}. We only had the model predictions available and extrapolated overlap from that, we did not have access to the models themselves, ground truth test set labels, nor dev set predictions for the models.

\subsubsection{Comparisons and Claims}

We gather data from GLUE papers regarding the accuracy tasks and manually label 119 comparisons and 57 claims of improvement (as denoted within a work by bolding of a new model's number and a claim of SOTA in the main text) across 14 papers (selected as being at or above the BERT score on the GLUE leaderboard with an accompanying publication). For each paper we examine if a specific comparison is made against a baseline that isn't claiming state of the art performance. For example, the STILTs approach~\citep{phang2018sentence} makes comparisons against non-SOTA baselines, which we add to our labeling scheme but filter out when fitting regressions to likely SOTA improvements. We mark this as \textbf{SOTA Comparison} = N. For claims of SOTA improvement, we examine this as some textual basis for the claim (e.g., ``we drive state of the art performance on GLUE'') coupled with bolding of values in a table reporting baselines against the model under test. We mark datapoints as \textbf{Claim of Improvement} = Y if they are an improvement claim. We mark effect size as the improvement from the best previous baseline (the current SOTA) on the test set on a per-dataset basis. We note that in several cases, worse results on the new model were bolded. We treated this as no claim of improvement. If results were not bolded but still higher for the new model we also treated this as no claim for improvement. 

\subsection{Regression-based approach to modeling power and MDEs} \label{app:regressions}

\subsubsection{Predicting overlap}
\label{app:overlap}

There are several versions of McNemar's test, each with their own unique method for calculating power, sample size, or minimum effect size. See, for example, discussions in \citet{schlesselman1982case}, \citet{duffy1984asymptotic} \citet{suissa19912}, \citet{connett1987sample}, \citet{fagerland2013mcnemar}, and \citet{lachenbruch1992sample}.

The methods for calculating sample size or power by \citet{connett1987sample,schlesselman1982case,suissa19912} require making an assumption about the odds ratio $\Phi = p_{10}/p_{01}$ as well as an estimate of the fraction of discordant pairs (disagreements between two models).

\citet{fagerland2013mcnemar} suggest that the exact unconditional version of the test by \citet{suissa19912} has desirable properties.
Thus, we use the implementation of the power calculations for this test from the \url{https://github.com/ekstroem/MESS} package. 

How do we make an assumption about the odds ratio and fraction of discordant pairs? We first fit an OLS regression to the existing models on the GLUE leaderboard for all binary choice accuracy tasks using the aforementioned predictions provided by the leaderboard creators and individual authors of models,
\begin{equation}
\begin{split}
    \text{overlap}_i = \beta_0 &+ \beta_1 \text{min\_acc}_i + \beta_2 \text{acc\_diff}_i,
\end{split}
\end{equation}
for all $i$ that are a pairwise comparison between any two models, $\text{min\_acc}_i$ is the minimum accuracy between the two models under comparison, $\text{acc\_diff}_i$ is the gap between the two models, and $\text{overlap}_i$ is the fraction of overlapping predictions. We end up with the model shown in Table \ref{tab:app_ols_glue}.

\begin{table*}
\begin{center}
\begin{tabular}{lclc}
\toprule
\textbf{Dep. Variable:}    &        y         & \textbf{  R-squared:         } &     0.966   \\
\textbf{Model:}            &       OLS        & \textbf{  Adj. R-squared:    } &     0.966   \\
\textbf{Method:}           &  Least Squares   & \textbf{  F-statistic:       } &     3820.   \\
\textbf{Date:}             & Thu, 14 May 2020 & \textbf{  Prob (F-statistic):} & 3.62e-197   \\
\textbf{Time:}             &     07:03:28     & \textbf{  Log-Likelihood:    } &    818.14   \\
\textbf{No. Observations:} &         270      & \textbf{  AIC:               } &    -1630.   \\
\textbf{Df Residuals:}     &         267      & \textbf{  BIC:               } &    -1619.   \\
\textbf{Df Model:}         &           2      & \textbf{                     } &             \\
\bottomrule
\end{tabular}
\begin{tabular}{lcccccc}
               & \textbf{coef} & \textbf{std err} & \textbf{t} & \textbf{P$> |$t$|$} & \textbf{[0.025} & \textbf{0.975]}  \\
\midrule
\textbf{const} &       0.4142  &        0.019     &    21.694  &         0.000        &        0.377    &        0.452     \\
\textbf{min\_acc}    &       0.5819  &        0.021     &    27.999  &         0.000        &        0.541    &        0.623     \\
\textbf{acc\_diff}    &      -0.4662  &        0.028     &   -16.625  &         0.000        &       -0.521    &       -0.411     \\
\bottomrule
\end{tabular}
\begin{tabular}{lclc}
\textbf{Omnibus:}       &  6.121 & \textbf{  Durbin-Watson:     } &    1.040  \\
\textbf{Prob(Omnibus):} &  0.047 & \textbf{  Jarque-Bera (JB):  } &    8.647  \\
\textbf{Skew:}          & -0.108 & \textbf{  Prob(JB):          } &   0.0133  \\
\textbf{Kurtosis:}      &  3.850 & \textbf{  Cond. No.          } &     71.5  \\
\bottomrule
\end{tabular}
\caption{OLS Regression Results for predicting GLUE model overlap from baseline accuracy and effect size.}
\label{tab:app_ols_glue}
\end{center}
\end{table*}

\begin{table*}
\begin{center}
\begin{tabular}{lclc}
\toprule
\textbf{Dep. Variable:}    &        y         & \textbf{  R-squared:         } &     0.944   \\
\textbf{Model:}            &       OLS        & \textbf{  Adj. R-squared:    } &     0.933   \\
\textbf{Method:}           &  Least Squares   & \textbf{  F-statistic:       } &     91.87   \\
\textbf{Date:}             & Tue, 26 May 2020 & \textbf{  Prob (F-statistic):} &  1.37e-07   \\
\textbf{Time:}             &     06:05:23     & \textbf{  Log-Likelihood:    } &    36.368   \\
\textbf{No. Observations:} &          14      & \textbf{  AIC:               } &    -66.74   \\
\textbf{Df Residuals:}     &          11      & \textbf{  BIC:               } &    -64.82   \\
\textbf{Df Model:}         &           2      & \textbf{                     } &             \\
\bottomrule
\end{tabular}
\begin{tabular}{lcccccc}
               & \textbf{coef} & \textbf{std err} & \textbf{t} & \textbf{P$> |$t$|$} & \textbf{[0.025} & \textbf{0.975]}  \\
\midrule
\textbf{const} &       0.4339  &        0.091     &     4.786  &         0.001        &        0.234    &        0.633     \\
\textbf{min\_acc}    &       0.5932  &        0.101     &     5.874  &         0.000        &        0.371    &        0.816     \\
\textbf{acc\_diff}    &      -1.2849  &        0.588     &    -2.186  &         0.051        &       -2.578    &        0.009     \\
\bottomrule
\end{tabular}
\begin{tabular}{lclc}
\textbf{Omnibus:}       &  0.299 & \textbf{  Durbin-Watson:     } &    2.022  \\
\textbf{Prob(Omnibus):} &  0.861 & \textbf{  Jarque-Bera (JB):  } &    0.163  \\
\textbf{Skew:}          &  0.214 & \textbf{  Prob(JB):          } &    0.922  \\
\textbf{Kurtosis:}      &  2.691 & \textbf{  Cond. No.          } &     140.  \\
\bottomrule
\end{tabular}
\caption{OLS Regression Results for predicting SQuAD 2.0 model overlap.}
\label{tab:app_ols_squad}
\end{center}
\end{table*}

We note that outcomes are biased toward a higher range of accuracy values and may not be a perfect prior. However, this does give us a fairly good linear fit for top-of-the-leaderboard results. We then can predict the expected overlap for a given model as:
\begin{equation}
\begin{split}
    \text{exp\_overlap} = &0.41 + 0.58 \cdot \text{ min\_acc} \\ &- 0.47 \cdot \text{exp\_acc\_dif}
\end{split}
\end{equation}

Note now we can make an assumption on the expected fraction of discordant values and the odds ratio, the latter being:
\begin{equation}
    \Phi = \frac{1 - \text{exp\_overlap} + \text{exp\_acc\_diff}}{1 - \text{exp\_overlap} - \text{exp\_acc\_diff}}
\end{equation}

This is all that is necessary for McNemar's test and thus we can then simply solve for the minimum expect treatment effect for the given sample size of the dataset and a power of $80\%$. Note that for QQP we use the normal approximation rather than exact unconditional test as the large sample size makes the exact test intractable. See \citet{duffy1984asymptotic}.

\begin{figure}
    \centering
    \includegraphics[width=.35\textwidth]{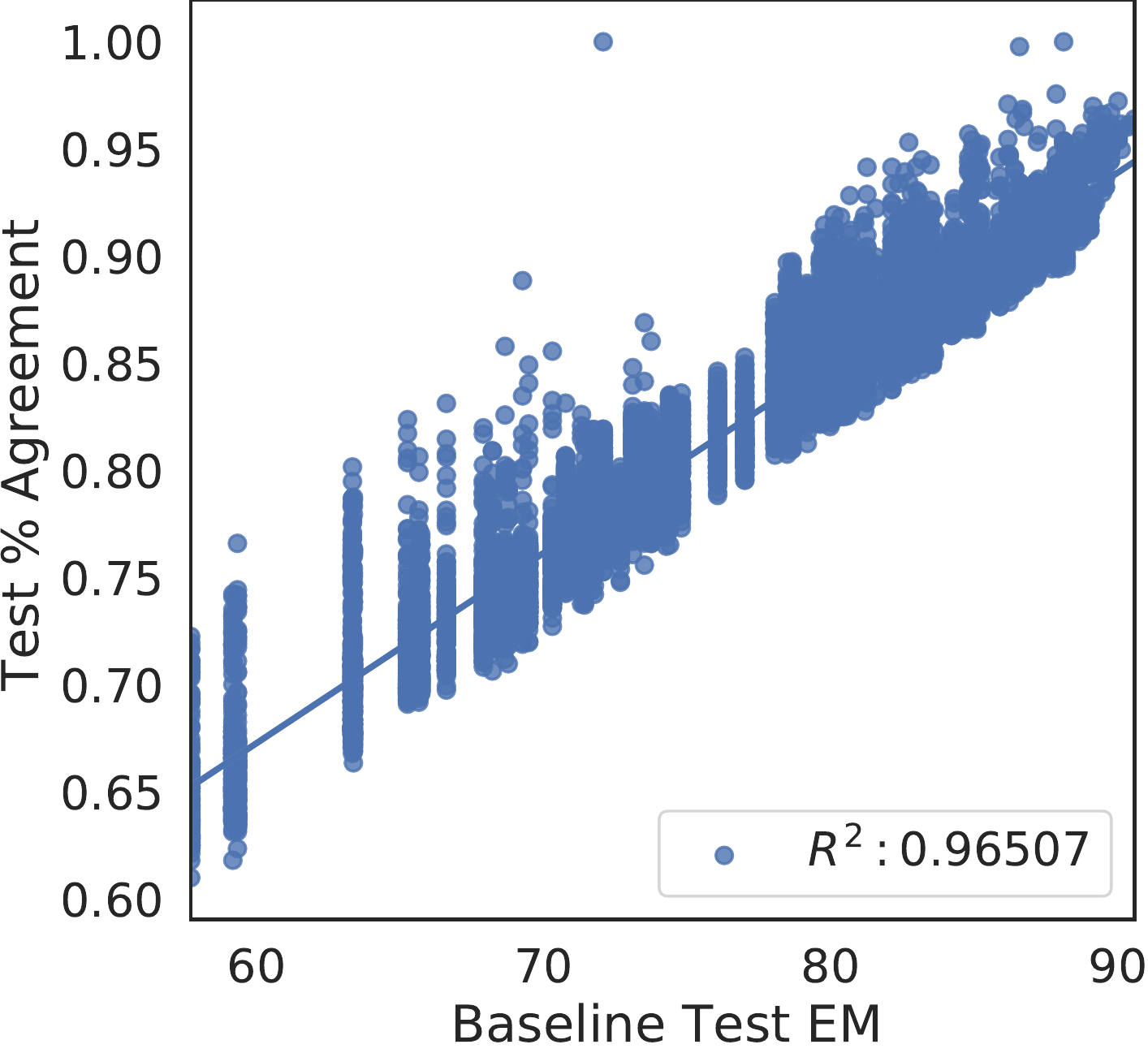}
    \includegraphics[width=.35\textwidth]{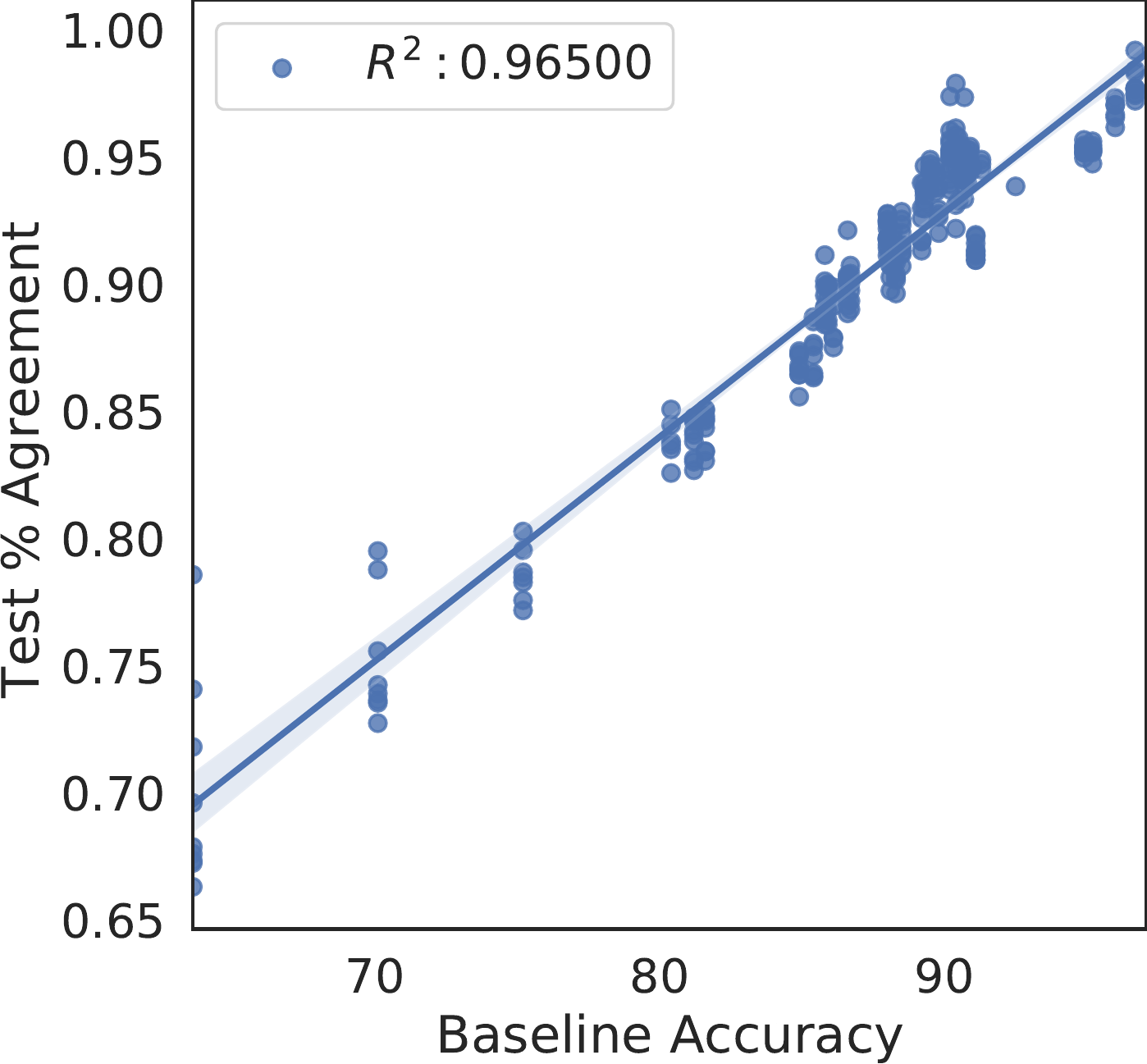}
    \caption{SQuAD 2.0 (top) and GLUE (bottom) \% agreement of new model vs. the accuracy of the baseline in the comparison (assuming improvement in the new model).}
    \label{fig:squadbaselinevagreement}
\end{figure}

We fit such a regression to GLUE tasks and achieve an $R^2$ of $0.97$. Repeating this for SQuAD 2.0, we get an $R^2$ of $0.94$, with fit shown in Table \ref{tab:app_ols_squad}. See Figure~\ref{fig:squadbaselinevagreement} for a plot indicating the level of agreement plotted against baseline accuracy. See also additional model comparisons for overlap in Appendix~\ref{app:modeloverlapplots}.

\subsubsection{Predicting Effect Size}
\label{app:effectsize}

A similar regression can be run to predict the expected effect size given the baseline accuracy: how much do models typically improve given the current SOTA. To fit an OLS regression predicting this value, we gather data from GLUE papers regarding the accuracy tasks and manually label 119 comparisons and 57 claims of improvement (as denoted within a work by bolding of a new model's number and a claim of SOTA in the main text) across 14 papers (selected as being at or above the BERT score on the GLUE leaderboard with an accompanying publication).  
We fit the regression:
\begin{equation}
    \widehat{\Delta}_i = \beta_0 + \beta_1 \text{baseline}_i + \hat{\beta}_2 \text{task}_i,
\end{equation}
\noindent to see how predictable the expected effect size is, where $\widehat{\Delta}_i$ is the predicted effect size, $\text{baseline}_i$ is the baseline model's accuracy, and $\text{task}_i$ is a categorical variable (in the regression this ends up being a set of dummy variables for each category so we denote $\hat{\beta}$ to emphasize this). Note that for SQuAD 2.0, we use a separate regression without the task variable since it is a single-task leaderboard.

We achieve an $R^2 = 0.69$ which is not a perfect fit, but still provides a prior on likely effect size. Similarly, we achieve an $R^2=.67$ when fitting a regression to SOTA improvements on the SQuAD 2.0 leaderboard (selected as being a significant improvement in time-ordered submissions).

\begin{table}[]
\small
\centering 
\begin{tabular}{@{\extracolsep{5pt}}lD{.}{.}{-3} } 
\\[-1.8ex]\hline 
\hline \\[-1.8ex] 
 & \multicolumn{1}{c}{\textit{Dependent variable:}} \\ 
\cline{2-2} 
\\[-1.8ex] & \multicolumn{1}{c}{effect.size} \\ 
\hline \\[-1.8ex] 
 Previous.Best & -0.264^{***} \\ 
  & (0.032) \\ 
  & \\ 
 TaskMNLI-mm & 0.150 \\ 
  & (0.621) \\ 
  & \\ 
 TaskMRPC & 0.023 \\ 
  & (0.622) \\ 
  & \\ 
 TaskQNLI & 2.139^{***} \\ 
  & (0.639) \\ 
  & \\ 
 TaskQQP & -0.195 \\ 
  & (0.719) \\ 
  & \\ 
 TaskRTE & 1.018 \\ 
  & (0.628) \\ 
  & \\ 
 TaskSST-2 & 1.536^{**} \\ 
  & (0.686) \\ 
  & \\ 
 TaskWNLI & -0.520 \\ 
  & (0.789) \\ 
  & \\ 
 Constant & 24.342^{***} \\ 
  & (2.837) \\ 
  & \\ 
\hline \\[-1.8ex] 
Observations & \multicolumn{1}{c}{61} \\ 
R$^{2}$ & \multicolumn{1}{c}{0.690} \\ 
Adjusted R$^{2}$ & \multicolumn{1}{c}{0.642} \\ 
Residual Std. Error & \multicolumn{1}{c}{1.309 (df = 52)} \\ 
F Statistic & \multicolumn{1}{c}{14.455$^{***}$ (df = 8; 52)} \\ 
\hline 
\hline \\[-1.8ex] 
\textit{Note:}  & \multicolumn{1}{r}{$^{*}$p$<$0.1; $^{**}$p$<$0.05; $^{***}$p$<$0.01} \\ 
\end{tabular} 
\caption{OLS regression for predicting effect size for GLUE tasks.} 
\label{tab:OLSeffectsizeglue} 
\end{table} 

\begin{table*}[]
\begin{center}
\begin{tabular}{lclc}
\toprule
\textbf{Dep. Variable:}    &        y         & \textbf{  R-squared:         } &     0.672   \\
\textbf{Model:}            &       OLS        & \textbf{  Adj. R-squared:    } &     0.644   \\
\textbf{Method:}           &  Least Squares   & \textbf{  F-statistic:       } &     24.55   \\
\textbf{Date:}             & Tue, 26 May 2020 & \textbf{  Prob (F-statistic):} &  0.000334   \\
\textbf{Time:}             &     06:05:23     & \textbf{  Log-Likelihood:    } &    45.711   \\
\textbf{No. Observations:} &          14      & \textbf{  AIC:               } &    -87.42   \\
\textbf{Df Residuals:}     &          12      & \textbf{  BIC:               } &    -86.14   \\
\textbf{Df Model:}         &           1      & \textbf{                     } &             \\
\bottomrule
\end{tabular}
\begin{tabular}{lcccccc}
               & \textbf{coef} & \textbf{std err} & \textbf{t} & \textbf{P$> |$t$|$} & \textbf{[0.025} & \textbf{0.975]}  \\
\midrule
\textbf{const} &       0.1331  &        0.023     &     5.910  &         0.000        &        0.084    &        0.182     \\
\textbf{x1}    &      -0.1408  &        0.028     &    -4.955  &         0.000        &       -0.203    &       -0.079     \\
\bottomrule
\end{tabular}
\begin{tabular}{lclc}
\textbf{Omnibus:}       & 19.911 & \textbf{  Durbin-Watson:     } &    2.643  \\
\textbf{Prob(Omnibus):} &  0.000 & \textbf{  Jarque-Bera (JB):  } &   18.487  \\
\textbf{Skew:}          &  1.995 & \textbf{  Prob(JB):          } & 9.68e-05  \\
\textbf{Kurtosis:}      &  6.971 & \textbf{  Cond. No.          } &     17.3  \\
\bottomrule
\end{tabular}
\caption{OLS Regression Results for predicting effect size from baseline accuracy for SQuAD 2.0 improvements.
    \label{tab:olseffectsizesquad}
}
\end{center}
\end{table*}

\begin{figure}
    \centering
    \includegraphics[width=.4\textwidth]{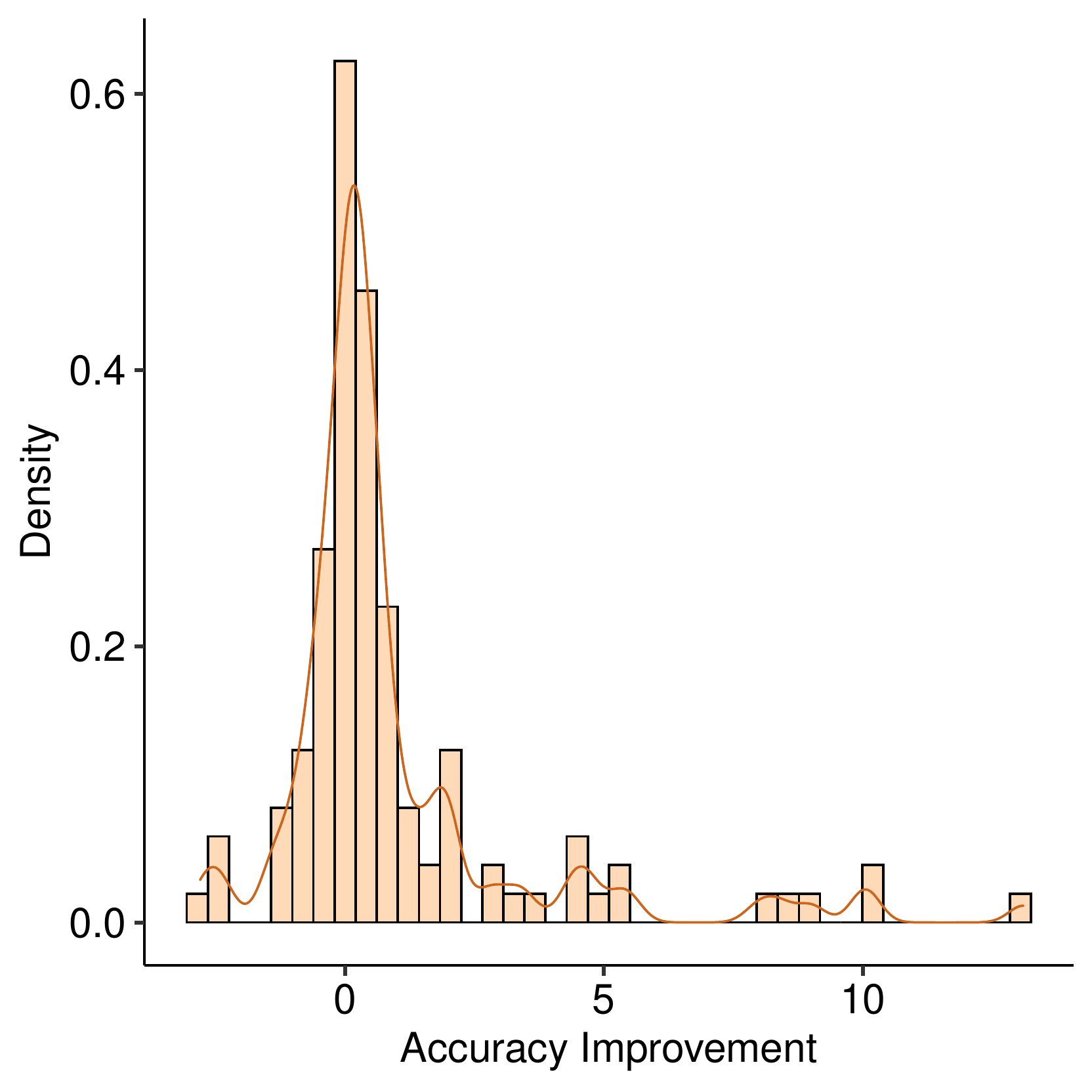}
    \caption{The reported difference from the best performing new model to the best performing baseline in accuracy across all accuracy datasets in the GLUE Benchmark. Note: unlike Table~\ref{ref:pvaluedistro}, we do not limit these to claims of improvement, but only to papers which introduce a new model and compare against some baseline. Mean: +0.959 Std.Err.: 0.23}
    \label{fig:effect_distribution}
\end{figure}

\begin{figure}
    \centering
    \includegraphics[width=.45\textwidth]{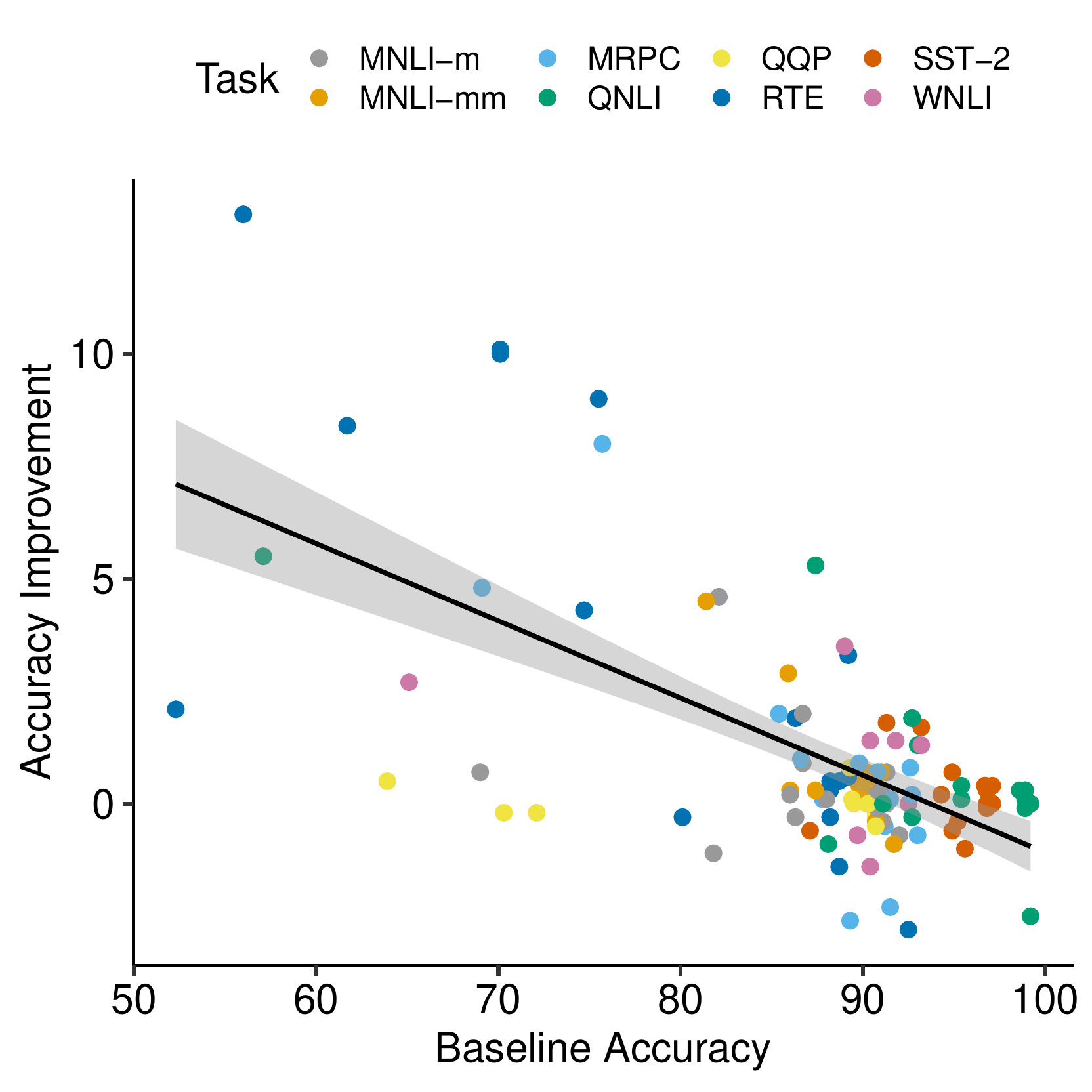}
    \caption{The effect size given the baseline model accuracy observed across GLUE tasks. As the baseline model moves toward the range of current GLUE submissions, reported model gains decrease toward 0. Fitting a regression yields an $R^2 = 0.69$.}
    \label{fig:effect_distribution_per_task}
\end{figure}

See Table~\ref{tab:OLSeffectsizeglue} 
and Table~\ref{tab:olseffectsizesquad} 
for regression coefficients and model fits. Figure~\ref{fig:effect_distribution_per_task} shows the per-task distribution of effect sizes against baseline accuracies in GLUE papers for SOTA improvements. Figure~\ref{fig:effect_distribution} shows the effect size distribution as a histogram.

\subsubsection{Caveats for Regression-based Approach} Fitting a regression to predict overlap between a baseline and a new model has a good linear fit. However, this may not be the case for every dataset. Additionally, predicting effect sizes via a linear fit is not a perfect prior. The measurements of power in this case are meant to simulate estimating power before running evaluation on a test set, as running power analysis using only the observed effect may lead to the issues of post-hoc power estimation. 

\subsection{No Prior Approach \citep{lachenbruch1992sample}} 
\label{app:lachenbruch}

What do you do if there is no prior data available (as in a new task) and so you cannot make assumptions about discordant pairs or odds ratio? \citet{lachenbruch1992sample} discusses this exact problem in the context of clinical trials, and proposes an alternative method based on the work of \cite{connett1987sample} which allows you to make assumptions about potential marginal probabilities, providing a midpoint value, as well as an upper and lower bound. We use an implementation of this from: \url{https://rdrr.io/rforge/biostatUZH/man/sampleSizeMcNemar.html} and solve for the expected accuracy minimum given a fixed dataset sample size and baseline accuracy for each of the lower bound, midpoint, and upper bound. In practice, we find the \citet{lachenbruch1992sample} prior to be very close to the values we obtain from the above regression (see Table \ref{tab:glueexpanded}). Importantly this method requires no assumptions and is meant to give an idea for whether it is worth pursuing a study for the given size of the test set.

\subsection{Extended Results}
\label{app:glueexpandedres}

Table~\ref{tab:glueexpanded} contains additional MDE estimates using a two-sample proportion test as in Appendix~\ref{app:binomial}, the \citet{lachenbruch1992sample} methodology. We also provide the standard errors and $n$ for each average effect size, the OLS regression predicting the next effect size for a new SOTA $\widehat{\Delta}$, and the current difference from SOTA and next on the leaderboard. We note that MDE calculations are roughly similar except for the upper and lower bounds provided in the \citet{lachenbruch1992sample} calculation. We also note that predicted SOTA results are far lower than past averages since the average includes early large results like those of \citet{devlin2018bert}. We can see that in some cases the predicted effect size is even smaller than the lowest bound MDE and we may wish to consider the usefulness of further comparisons on individual datasets in such cases.

\begin{table*}[]
    \centering \small
        \resizebox{\textwidth}{!}{
    \begin{tabular}{|c|c|c|c|c|c|c|c|c|}
         \hline
         Dataset & Size & SOTA & MDE Binomial & MDE  \citep{lachenbruch1992sample} & MDE regression & $\widehat{\Delta}$ & $|\Delta|$ (std.err.,n) & $\Delta_{SOTA}$\\
         \hline
        WNLI&147&94.5\%&+5.38\%&+5.42\%(5.36\%, 5.45\%)&+5.26\% & -1.17\% & 1.72 (0.917, 4) & 0.0\%\\         
      MRPC&1725& 92.0\%  &+2.40\% & +1.91\% (0.45\%, 2.48\%) & +1.62\% & +0.03\%& +0.625 (0.234, 8)& +0.6\%\\
      SST-2& 1821 & 97.2\% &+1.34\% & +1.10\% (0.43\%,1.35\%) & +1.02\% &+0.18\%  & +0.571 (0.197, 7)& -0.3\%\\    
      RTE&3000 & 91.7\%&+1.89\%& +1.48\% (0.26\%, 1.96\%) & +1.23\% & +1.11\%& +3.89 (1.23, 10)&+0.8\%\\      
      QNLI&5463&97.5\%&+0.77\%& +0.60\% (0.14\%, 0.78\%) & +0.55\% & + 0.69\% & +1.31 (0.552, 9)& +0.9\%\\
      MNLI-m & 9796 & 91.6\% &+1.08\% & +0.82\% ( 0.08\%, 1.12\%) & +0.67\% & +0.12\% & + 0.97 (0.442, 10)&+0.2\%\\
      MNLI-mm & 9847& 91.3\%&+1.09\%& +0.84\% ( 0.08\%, 1.14\%)& +0.68\% & +0.34\% & + 1.29 (0.550, 8)&+0.3\%\\
      QQP &390965&91.0\%& +0.18\% & + 0.13\% ($8.45 \times 10^{-5}$\%, 0.19\%) & +0.11\% & +0.08\% & 0.36 (0.121, 5)& +0.1\%\\
         \hline
         SQuAD 2.0 & 8862&90.724\%& +1.18\%&+0.91\% (0.09\%, 1.23\%) & +0.556\%&+0.528\% & +2.23\% (0.431,14) $\dagger$& +0.146\%\\
      \hline
    \end{tabular}
    }
    \caption{The minimum detectable effect (MDE) for various datasets given the current top accuracy on the leaderboard on May 6th, 2020. See Appendix~\ref{app:mde_table} for expanded details. How to use this table? Suppose you are building a model to get SOTA on any of these datasets. If you don't have a reasonable expectation that your model will exceed the MDE, then it is not worth proceeding with the study on a dataset of this size and instead either more data should be collected or a different (larger) dataset used.
    MDE \citep{lachenbruch1992sample} provides a mid-point and upper/lower bound assumptions using the most conservative and generous estimates of model agreement. 
    MDE Binomial uses the binomial test as the assumed statistical test and calculates the MDE using the exact mechanism from Appendix~\ref{app:binomial}.
    See also discussion by \citet{arend2019statistical}.
    $\widehat{\Delta}$ is the expected effect by fitting a regression to all SOTA improvement claims found in reviewed papers. $|\Delta|$ (std.err., n) is the average improvement in surveyed papers that claimed SOTA and had a positive effect size reported for the dataset (with standard error and the number of papers in parentheses). $\dagger$ indicates that the SQuAD 2.0 average improvement was based on improvements to the SQuAD leaderboard, but weren't necessarily reported as improvements in a publication.
    $\Delta_{SOTA}$ is the gap between the SOTA model (ALBERT + DAAF + NAS) on GLUE and the next best model (ERNIE) -- this was not included in the regression.
    }
    \label{tab:glueexpanded}
\end{table*}

\begin{table*}[] \centering 
    \resizebox{\textwidth}{!}{

  \begin{tabular}{@{\extracolsep{5pt}}lD{.}{.}{-3} D{.}{.}{-3} D{.}{.}{-3} D{.}{.}{-3} D{.}{.}{-3} D{.}{.}{-3} D{.}{.}{-3} } 
\\[-1.8ex]\hline 
\hline \\[-1.8ex] 
Statistic & \multicolumn{1}{c}{N} & \multicolumn{1}{c}{Mean} & \multicolumn{1}{c}{St. Dev.} & \multicolumn{1}{c}{Min} & \multicolumn{1}{c}{Pctl(25)} & \multicolumn{1}{c}{Pctl(75)} & \multicolumn{1}{c}{Max} \\ 
\hline \\[-1.8ex] 
Power & 57 & 0.698 & 0.352 & 0.034 & 0.407 & 1.000 & 1.000 \\ 
P & 57 & 0.220 & 0.283 & 0.000 & 0.00000 & 0.348 & 1.000 \\ 
\hline
Statistic & \multicolumn{1}{c}{N} & \multicolumn{1}{c}{Percentage} & \multicolumn{5}{c}{-} \\ 
\hline
\% Powered & 57 & 0.456 \% & - & - & - & - & - \\ 
\% Significant & 57 & 0.509 \% & - & - & - & - & - \\ 
\% significant and Powered & 57 & 0.368\% & - & - & - & - & - \\ 
\hline \\[-1.8ex] 
\end{tabular} 
}
  \caption{
  We examine the claims of SOTA improvement in surveyed GLUE papers and use a leave-one-out regression-based estimate of effect size and overlap to simulate how many authors would have found their study to be well-powered. We also examine how many of the observed effects were likely significant based on predicted model overlap. We note that if we use the \emph{observed} effect in a post-hoc analysis, the proportion of studies falling below the MDE is even higher.
      \label{ref:pvaluedistro}
} 
\end{table*} 

\begin{figure}
    \centering
    \includegraphics[width=.5\textwidth]{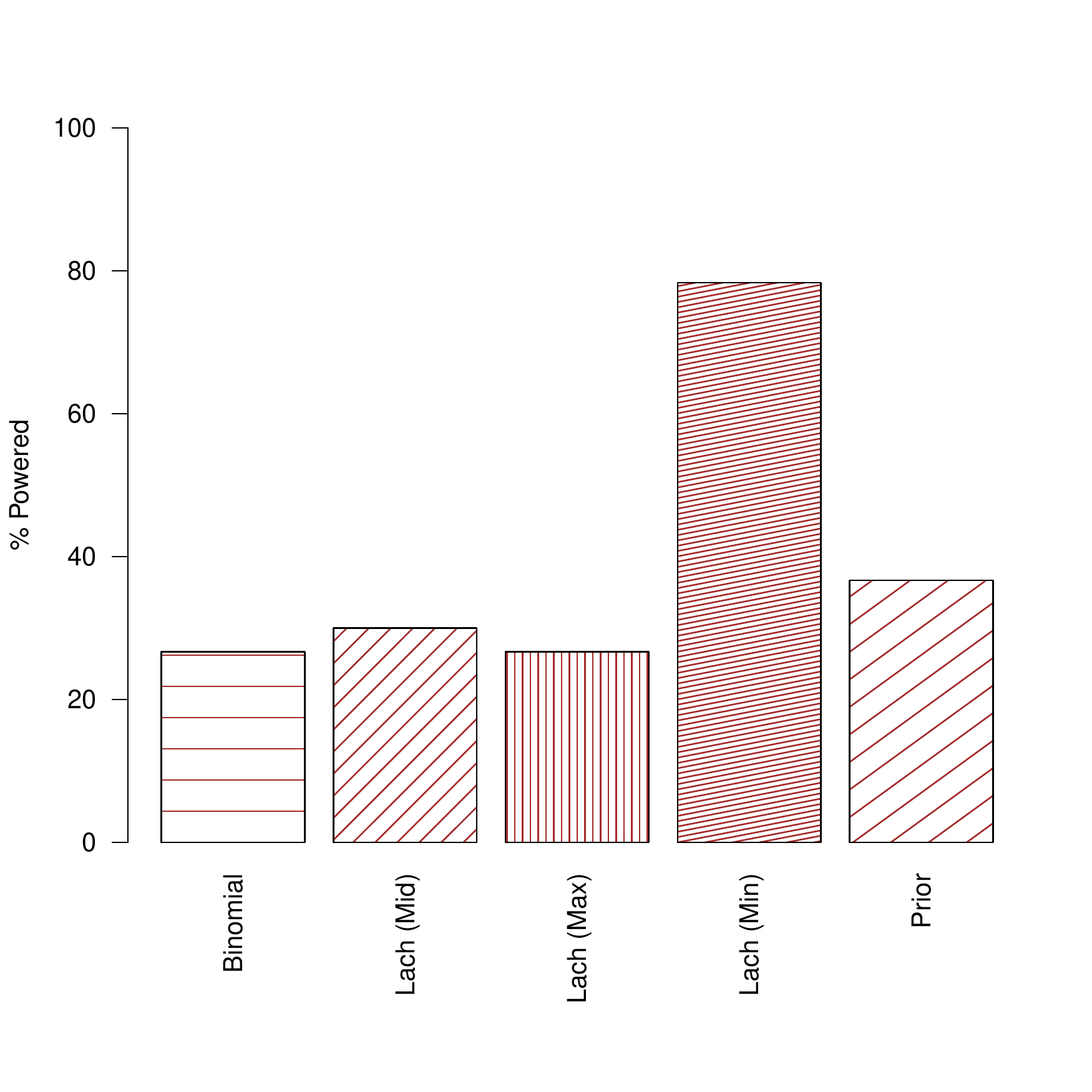}
    \caption{Of the claims of improvement over a given baseline (indicated in text and via bolded values in tables) across 14 papers on the GLUE leaderboard (also seen in Table~\ref{ref:pvaluedistro}). We find only 26.7\% of observed effects met the MDE to the binomial power calculation, 30\% met the MDE according to the midpoint calculation of \cite{lachenbruch1992sample}, 26.7\% met the MDE when using the upper bound from the \cite{lachenbruch1992sample} calculation, 78.3\% met the MDE when using the most generous (unlikely) assumptions for power according to the MDE \cite{lachenbruch1992sample} calculation, and 36.7\% met the MDE when using the regression-fitted prior of model overlap. Note: this assumes the true population effect \emph{is} the test set effect size. While this is post-hoc power analysis, we felt it may be useful to consider in the context that for a given model comparison on a given test set there is no variance and thus post-hoc power analysis is acceptable. However, for claims that include the entire data distribution this no longer holds and we refer back to the main text.}
    \label{fig:percentage_powered}
\end{figure}

\subsection{Calculating Power or Sample Size with Binomial Test}
\label{app:binomial}

If we assume that samples are \emph{unpaired} -- the new model and baseline evaluation samples are drawn from the same data distribution but aren't necessarily the same samples -- we can use a binomial test for significance. 

In this case, we assume that we have two models and each draw brings a 1 if the model is correct or 0 if incorrect.  We would like to use the two-sample proportion test, and have two binomial distributions with $p_1$ and $p_2$ as the mean probabilities. Our null hypothesis is $H_0 : p_1 = p_2$. We have an alternative hypothesis (two sided) is $H_1 : p_1 \ne p_2$. Note, in R we can use the function \textbf{power.prop.test()} to calculate power, the MDE, or the sample size of the tests. See also a tutorial here: \url{https://imai.fas.harvard.edu/teaching/files/Handout9.pdf}. 

\section{Additional Metrics}
\label{app:metrics}

In this appendix, we provide guidance on how we might apply power analysis to metrics beyond what is covered in the main paper.

\paragraph{Recall, Precision, F1, Matthew's correlation:}

While accuracy is the most commonly used metric in the GLUE benchmark, other tasks make use of other metrics such as F1 and Matthew's correlation. F1 is particularly relevant in cases of binary classification where there is strong class imbalance, such that even the baseline of predicting the most common class will achieve high accuracy. 

If we have good prior information, we can use an approach akin to that recommended for accuracy, but replacing McNemar's test with a randomization test (as used for machine translation, see \S\ref{sec:mt} in main paper). In particular, given an evaluation on paired data (as is the case for all benchmark datasets), one can test for a significant difference between models in terms of F1 (or any other metric) using a randomization test. That is, on each iteration, we randomize the assignment of which model each prediction came from for every instance with probability 0.5, and compute the resulting overall difference in F1. Repeating this thousands of times gives us the null distribution, and we can then check to see whether the observe difference in F1 is in the tails of this distribution, which can thereby be converted into a $p$-value (see \citet{dror.2018} for more details).

Because F1 (and related metrics) cannot be represented as a simple sum over individual instances, in order to completely specify a hypothetical data generating process, we need to assume values for all cells in the confusion matrix, per class. That is for each class we would need to assume values for the cells as shown in Table \ref{tab:contingency_f1}, where the relevant distribution of predictions are for the instances with the corresponding label, and the values for each class sum to one.

\begin{table}[!ht]
\begin{center}
\small
\begin{tabular}{l|cc}
        & M1 negative & M1 positive\\\hline
M2 negative & $p(\textrm{both neg.})$ & $p(\textrm{only M1 pos.})$\\ 
M2 positive & $p(\textrm{only M2 pos.})$ & $p(\textrm{both pos.})$\\ 
\end{tabular}
\end{center}
    \caption{A contingency table representing the distribution of possible outcomes for two models (M1 and M2) on the instances of a single class of labels. The cells of this table should sum to 1.0 for each class}
    \label{tab:contingency_f1}
\end{table}

In addition, we need to assume the true distribution of labels in the data distribution of interest, $p(c)$ for $c$ in $\{1,\ldots,C\}$. Given these assumptions, we could then simulate an arbitrary number of datasets from this process. For each instance, we would first sample a true label $(c)$, and then sample the model predictions from the corresponding contingency table. For each simulated dataset, we could then apply the randomization test (using thousands of randomizations). By repeating this process many times, we can directly estimate power for the corresponding assumptions and sample size $n$. 

This process is not particularly efficient, but can still be run relatively quickly on a laptop. The more difficult part is choosing good values for the necessary probabilities. However, such an approach can still be used to test for how sensitive power is to variations in assumptions. It is also possible to make simplifying assumptions, such as that the rate of false positives and false negatives will be the same across classes, or to estimate some parameters from training data, such as the underlying distribution of labels. The same technique can easily be extended to other metrics that depend on the contingency table, such as Matthew's correlation.

\section{Additional Details for the BLEU Scores Power Analysis}
\label{app:mt}

In this section, we provide further details for the machine translation (MT) data generation procedure as well as an analysis of how power varies for a range of values of $P_0$ and $b_0$, the parameters estimated from the empirical observations.

\subsection{Data Generation Procedure}
\label{app:mt_datagen}
Recall that using the randomization test to determine whether two MT systems are statistically different gives rise to the null distribution of differences in BLEU.\footnote{The bootstrap is another valid approach to testing for differences between models \citep{koehn.2004,graham.2014,dror.2018}, though note the concerns highlighted by \citet{riezler.2005}.}
If we had access to large amounts of parallel text, we could instead sample many subsets of real sentences and evaluate the difference between models on those subsets, which allow us to characterize the mean and variance of the difference in model performance. Such estimates could then be used to estimate power directly.
Because we do not have access to such data, however, we instead rely on the randomization approach, 
in which we run several thousand trials where the paired output translations for a subset of the test set samples are swapped.
In order to estimate power, we would like to be able to generate many datasets from a data generating procedure, which we can parameterize by various parameters, such as the difference between models.
Rather than generating raw text, however, and computing BLEU scores on that, we instead attempt to generate only the data necessary for the randomization test.
How can we do this?

In our case, the answer to this question lies in establishing a relationship between individual samples and the permuted set within each trial of the randomization test.
This relationship is as follows: \emph{the sum of individual changes to the difference in BLEU, from swapping single samples at a time, closely approximates the net change to the difference in BLEU, from swapping those samples all at once.}\footnote{Note that this does not directly solve the problem of computing BLEU at the sentence level \citep{chen.2014}, as it still mimicking the process of evaluating BLEU on a corpus.}
Let $S$ be the set of test set samples swapped during a single trial of the randomization test and $R_B(S)$ be the difference in BLEU between the paired outputs after swapping the examples in $S$.
$\Delta_B$ is the original difference in BLEU and $\delta_i$ is the change to the difference in BLEU from swapping test sample $i$ and leaving all other samples unswapped.
Then, we find that,
\begin{equation*}
    \sum_{i \in S}{\delta_i} \approx R_B(S) - \Delta_B
\end{equation*}

\begin{figure*}[!ht]
	\centering
	\begin{subfigure}[t]{0.49\textwidth}
		\centering
		\includegraphics[width=0.85\textwidth]{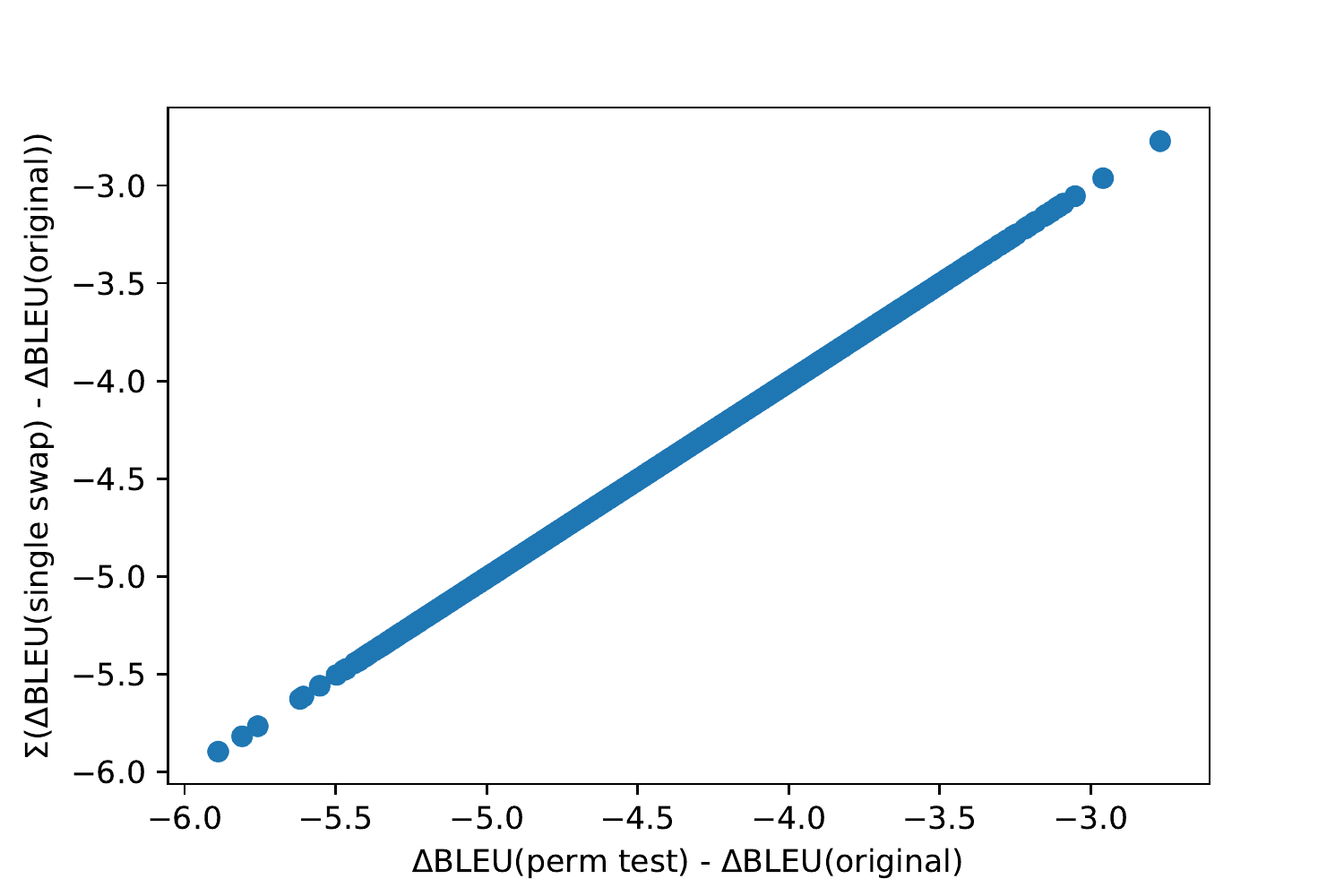}
		\caption{Model trained on WMT19 data versus model trained on WMT18 data, evaluated on the 2019 test set.}		
		\label{fig:mt_correlation}
	\end{subfigure}
	\hfill
	\begin{subfigure}[t]{0.49\textwidth}
		\centering
		\includegraphics[width=0.85\textwidth]{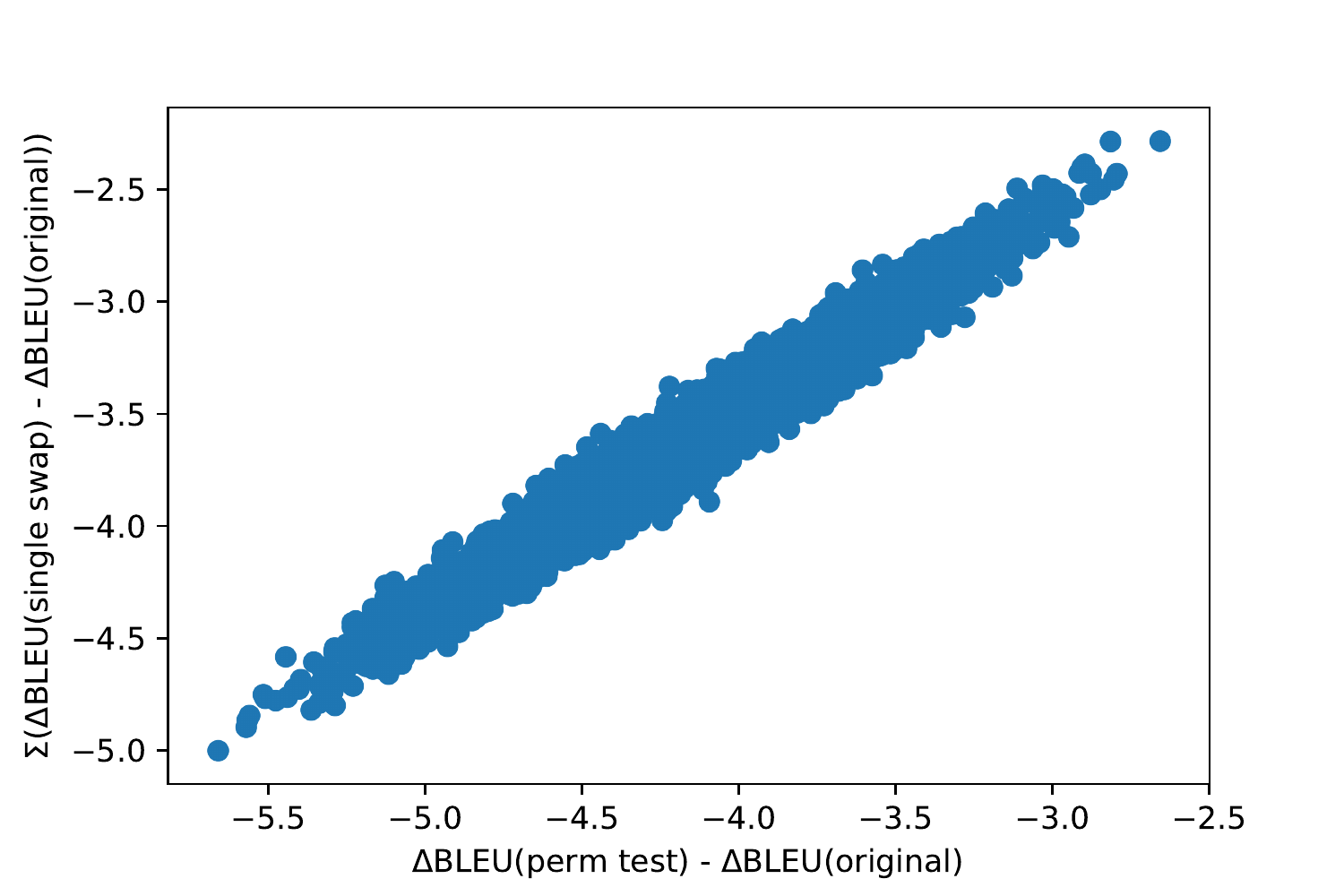}
		\caption{Model trained on WMT18 data versus model trained on WMT16 data, evaluated on the 2018 test set.}		
		\label{fig:mt_correlation_18v16}
	\end{subfigure}
	\caption{Correlation between individual changes to $\Delta_B$ and the net effect.} 
	\label{fig:mt_all_correlations}
\end{figure*}

\noindent This relationship is illustrated in Figure~\ref{fig:mt_all_correlations}:
Figure~\ref{fig:mt_correlation} shows the difference between two models evaluated on the 2019 test set, and Figure~\ref{fig:mt_correlation_18v16} shows the difference between a different pair of models evaluated on the 2018 test set.
We found the same relationship is true for the 2017 and 2016 test sets, as well. 

\begin{figure*}[t]
	\centering
	\begin{subfigure}[t]{0.49\textwidth}
		\centering
		\includegraphics[width=0.85\textwidth]{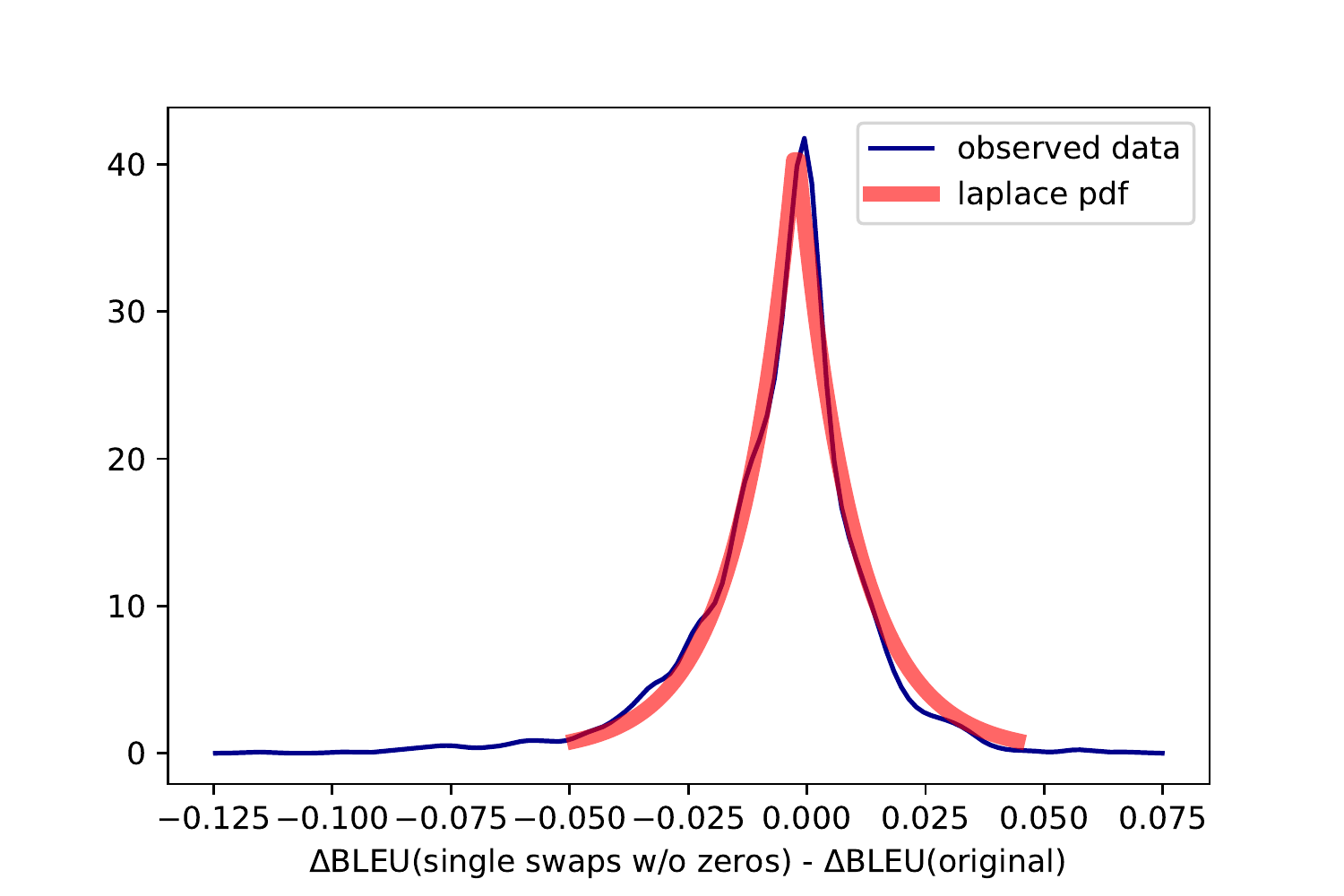}
		\caption{Model trained on WMT19 data versus model trained on WMT18 data, evaluated on the 2019 test set.}		
		\label{fig:mt_laplace}
	\end{subfigure}
	\hfill
	\begin{subfigure}[t]{0.49\textwidth}
		\centering
		\includegraphics[width=0.85\textwidth]{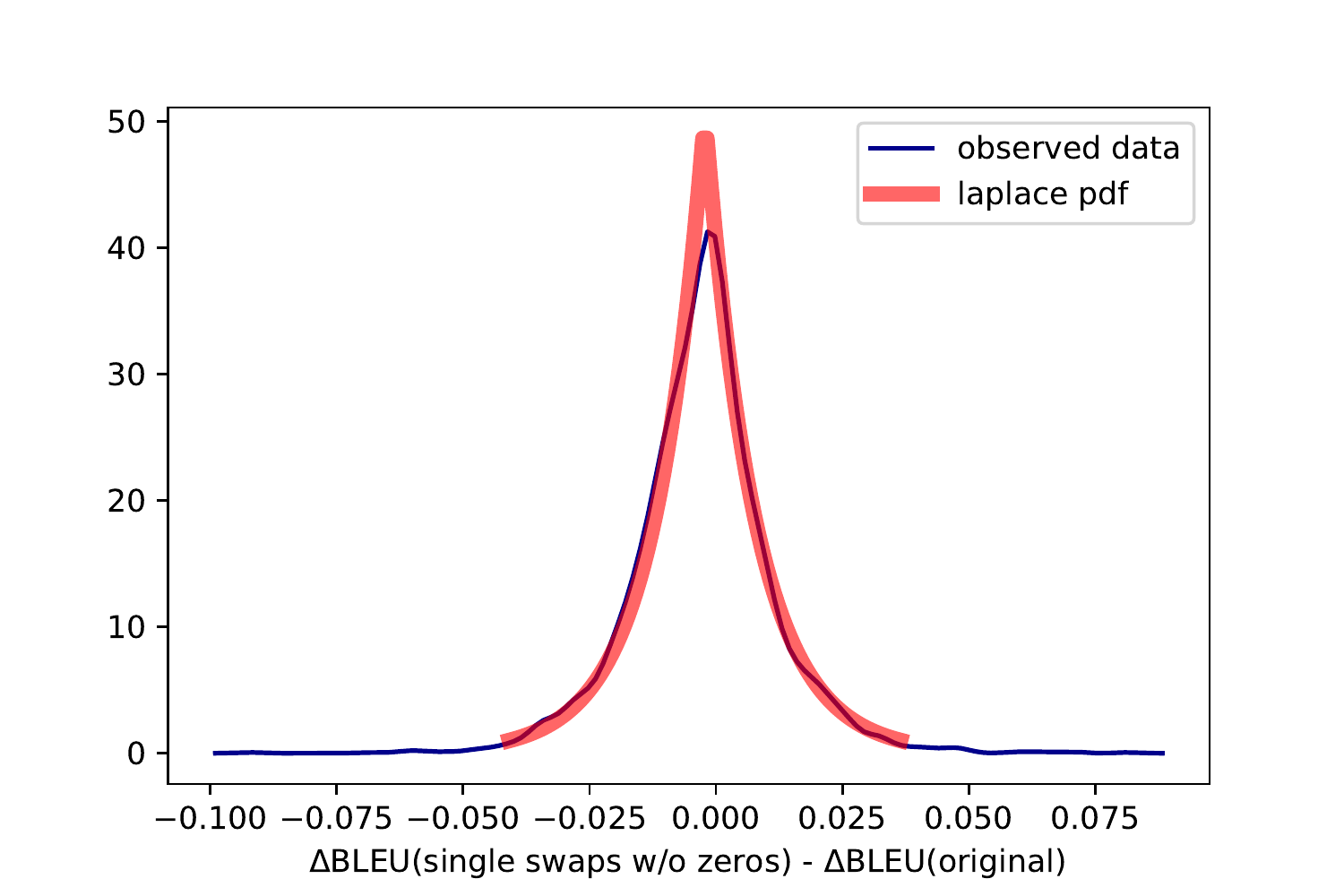}
		\caption{Model trained on WMT18 data versus model trained on WMT16 data, evaluated on the 2018 test set.}		
		\label{fig:mt_laplace_18v16}
	\end{subfigure}
	\caption{Fitting a Laplace distribution to individual non-zero effects.} 
	\label{fig:mt_all_laplace}
\end{figure*}

Now that we have established a relationship to closely approximate the outcome of each randomization trial, all that remains is to define a distribution from which the individual changes to the difference in BLEU can be sampled.
This distribution is a mixture of a Delta distribution at zero and a Laplace distribution.
The Delta distribution accounts for the proportion of samples ($P_0$) such that swapping any of them individually results in no change to the difference in BLEU, i.e. the effect is zero.
For the remaining samples, we fit a Laplace distribution, as shown in Figure~\ref{fig:mt_all_laplace}.
This Laplace is parametrized by two parameters: location ($\mu$) and scale ($b$).
By fitting this mixture to the individual effects computed from evaluating BLEU differences on many pairs of models, we discover that the variance parameter scales inversely proportional to the size of the dataset. Thus, we report an overall $b_0$ value for each dataset, such that $b_0$ = $b_k * n_k$, where $b_k$ is the Laplace scale parameter obtained from dataset $k$ containing $n_k$ samples.

For generating synthetic data, we need to specify $\mu$ and $b$, as well as $P_0$. However, because we want the effect of swapping half the non-zero samples from this distribution to equal the difference in BLEU between models, we only use the above fits to estimate $b_0$. We thus complete the generative process by assuming values for $\Delta_B$, $n$, $P_0$, $b_0$, and setting
$\mu = -2 \cdot \Delta_B / (n \cdot (1-P_0))$
such that the average effect of a random subset of $n/2$ instances is equal to $-\Delta_B$.
Table~\ref{tab:mt_params} in the main paper shows a range of observed values for $P_0$ and $b_0$.

\subsection{Variation in Power Estimates for a Range of Parameter Values}
\label{app:mt_power}

Now that we have defined the data generation procedure, and have estimates for the two parameters, $P_0$ and $b_0$, that are needed to simulate datasets, we can estimate power for a range of values for sample size $n$ and difference in BLEU $\Delta_B$, and see how these estimates vary as $P_0$ and $b_0$ change.
To provide a concrete example, suppose that we have two machine translation models that we expect will differ by $\Delta_B=$ 1 BLEU point. For a dataset of $n=$ 2,000 sentences, we assume that the models will perform equally for $P_0=0.2$, i.e. 20\% of sentences, and will assume a base scale parameter of $b_0=26$. To compute power, we would follow the process in Algorithm 1, with the following modifications. On each iteration, we would draw individual changes to the difference in BLEU from the distribution specified above, with $P_0 = 0.2$, $\Delta_B= 1$, $b_0=26$, and $n=2000$. For each such draw, we would apply the randomization test to compute a null distribution, using the sum of individual amounts as the total effect of flipping a random subset of pairs. Based on the null distribution, we compute if the difference is significant for this trial. Repeating this many times and observing the proportion of trials that are found to be significant gives us the approximate power.

\begin{figure*}[h]
\centering
\begin{subfigure}[h]{\textwidth}
    \centering
   \includegraphics[width=\linewidth]{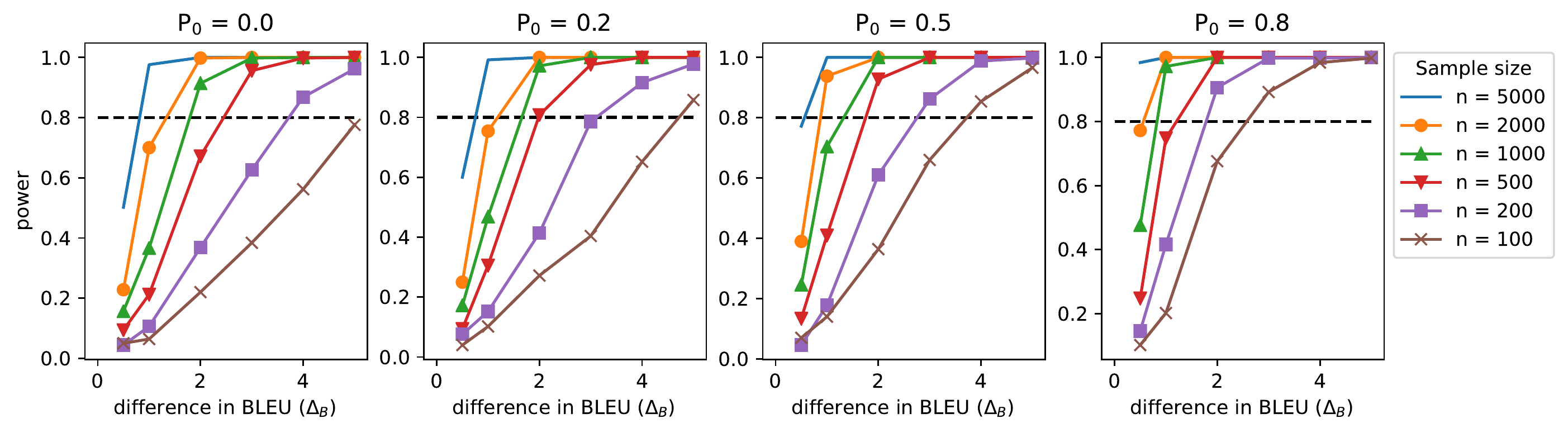}
   \label{fig:ref_mt_p0} 
\end{subfigure}

\begin{subfigure}[h]{\textwidth}
    \centering
   \includegraphics[width=\linewidth]{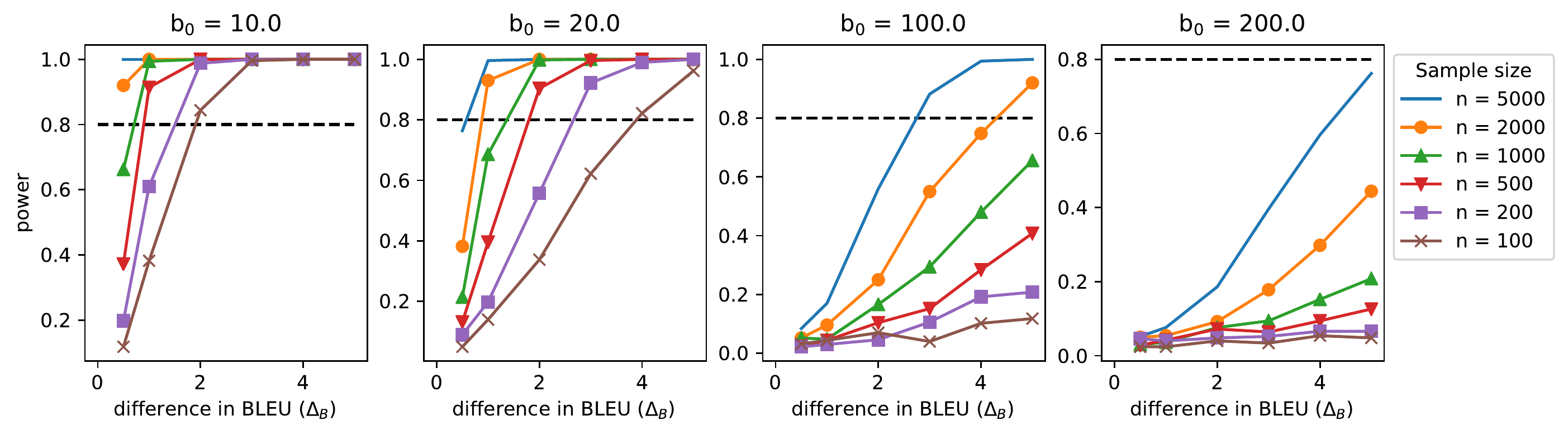}
   \label{fig:ref_mt_scale}
\end{subfigure}

\caption{Power Analysis for BLEU scores: Variation in estimates of power for different values of $P_0$ (top) and $b_0$ (bottom). For the top row, $b_0$ = 25.8, and for the bottom row, $P_0$ = 0.13.}
\label{fig:ref_mt_p0_and_scale}
\end{figure*}

Figure~\ref{fig:ref_mt_p0_and_scale} shows power for a range of values for $\Delta_B$, $n$, $P_0$ and $b_0$.
When $P_0$ is low, as is true for the observed data in Table~\ref{tab:mt_params}, effect sizes and sample sizes need to be larger in order for an experiment to be well-powered.
But as $P_0$ gets higher, a given effect size can be detected by a smaller sample size.
On the other hand, as $b_0$ increases and consequently the scale parameter $b$ for the Laplace grows, even large effect sizes cannot be detected by test sets containing 5,000 samples.

\section{Details of Human Evaluation Section}
\label{app:survey}

\subsection{Meta-analysis of human ratings for EMNLP 2019}

To assess the state of statistical power in a typical NLP study using human evaluation, we sampled papers from the mean EMNLP 2019 workshop that contained the phrase ``human eval''.  This first pass returned 117 papers, of which 86 had relevant human evaluations (in which models were compared), with the remainder either referencing human evaluation, or containing some other type of evaluation, such as comparing the agreement between automated metrics and human performance. Because some papers had more than one such evaluation, we had 97 experiments for analysis. Of these 51 were Likert experiments (as discussed in the main text), 38 were some form of direct model comparison, and 8 were other.

Significance testing was rare and was reported, in some form, in only 24\% of experiments. Bolding or starring the best results in a table was more common, occurring in 63\% of human rating experiments in our set. Whether bold results implies that the author is claiming a meaningful difference is not always clear. We did find one single case of authors performing a power analysis to estimate sample size among the papers we surveyed \citep{garbacea.2019}. However, because that paper did not involve a comparison of models to a baseline, it was not included in our analysis. In addition, we note that few details were provided, such that we were unable to ascertain precisely how the power analysis was done.

Because we chose to focus on ordinal ratings, we further annotated those in order to record the mean ratings and experimental characteristics (number of annotators, number of items, number of annotators per item), as well as all differences for all metrics between the model being proposed and the best performing baseline evaluated in the paper, as discussed in the main text.

\subsection{Human evaluation datasets}
\label{app:humandata}

For our analyses, we make use of the following datasets:
\begin{itemize}
    \item From \citet{hashimoto2019unifying} we use the evaluation data for Reddit, language modeling, and summarization. The data is available at \url{https://worksheets.codalab.org/worksheets/0x88644b5ee189402eb19d39d721d1005c}
    \item From \citet{dathathri.2020} we use the available ratings. The data is available at \url{https://github.com/uber-research/PPLM}
    \item For WMT19 (\url{http://statmt.org/wmt19/translation-task.html}), the data is available at \url{https://www.computing.dcu.ie/~ygraham/newstest2019-humaneval.tar.gz}
    \item For \citet{holtzman.2020}, we obtain the human evaluation data directly from the authors.
\end{itemize}

\subsection{Linear Mixed Effect Models}
\label{app:lmer}

To assess power in the human ratings framework, we used linear mixed effect models with random intercepts and slopes for worker and item, as in \citet{barr2013random}. Following best practices, we use the following structure, where $w$ is a particular worker and $i$ is a particular item. There are seven parameters, corresponding to the parameters needed for running a power analysis: fixed effects $\beta_{0}$ (the intercept) and $\beta_{1}$ (the model effect), and variance parameters for the worker intercept ($\sigma_{0w}$), the item intercept ($\sigma_{0i}$) and their respective slope variance parameters ($\sigma_{1w}$ and $\sigma_{1i}$). There is also a variance parameter for the overall error ($\sigma_{wi}$). We transform the Likert ratings to be on a [0, 1] scale and treat them as normally distributed (which we note is an imperfect assumption). 
We give fit parameters for these values, on a few datasets, in Tables \ref{table:fixef}, \ref{table:worker}, and \ref{table:item}.

\begin{align}
\begin{split}
Y_{wi} &= \beta_{0} + W_{0w} + I_{0i} \\ &+  (\beta_{1}+W_{1w} + I_{1i})X_{i} + e_{wi} 
\end{split} \\
I_{0i} &\sim N(0, \sigma_{0i}) \\
W_{0i} &\sim N(0, \sigma_{0w}) \\
I_{1i} &\sim N(0, \sigma_{1i}) \\
W_{1i} &\sim N(0, \sigma_{1w}) \\
e_{wi} &\sim N(0, \sigma_{wi})
\end{align}

For simplicity and convergence issues, we do not include a correlation parameter in the random effect structure.

To assess power, we use two possible variance settings derived from the model fits (``high variance'' and ``low variance'' settings, in the main text) and show these in Table \ref{table:lmer_recs}. We systematically vary the number of annotators (always assuming each annotator annotates each item, which is not always true in typical experiments), the number of items, and the effect size. We note that simulations can be customized to the planned analysis, including aspects such as how many items will be annotated by each annotator.

To compute power, we use each setting of the parameters to simulate 200 experiments and compute the proportion that detect a significant positive effect ($t > 1.96$). Significant effects in the opposite direction ($t < -1.96$) do not count as detections. Code for these model fits and simulations is included with the online materials. However, we note that these should be used as a starting point, rather than being blindly copied, as details may differ in each experimental setting.

\begin{table*}[]
\centering \small
\begin{tabular}{lrr}
  \hline
 Dataset & Number of Workers & Number of Items \\ 
  \hline
    \citet{hashimoto2019unifying} (LM) & 124 & 50 \\ 
  \citet{hashimoto2019unifying} (summarization) & 96 & 99 \\ 
\citet{hashimoto2019unifying} (Reddit) & 123 & 99 \\ 
  WMT19 & 176 & 1997 \\ 
  \citet{dathathri.2020}  & 15 & 1358 \\ 
  \citet{holtzman.2020} & 140 & 1399 \\ 
   \hline
\end{tabular}
\caption{Number of workers and items in each of our convenience sampled datasets.}
\end{table*}

\begin{table*}[]
\centering \small
\begin{tabular}{lrrrrrrrr}
  \hline
Dataset & $\hat{\hat{\beta_{0}}}$ & $\hat{\beta_{1}}$ & $\hat{\beta_{2}}$ & $\hat{\beta_{3}}$ & $\hat{\beta_{4}}$ & $\hat{\beta_{5}}$ & $\hat{\beta_{6}}$ & $\hat{\sigma}_{wi}$ \\ 
  \hline
\citet{hashimoto2019unifying} (LM) & 0.55 & -0.03 &  &  &  &  &  & 0.25 \\ 
  \citet{hashimoto2019unifying} (summarization) & 0.58 & 0.06 &  &  &  &  &  & 0.26\\
    \citet{hashimoto2019unifying} (Reddit) & 0.55 & 0.05 & 0.03 & 0.01 &  &  &  & 0.23 \\ 
  WMT19 & 0.86 & 0.04 &  &  &  &  &  & 0.12 \\ 
  \citet{dathathri.2020}  & 0.62 & 0.04 & -0.05 & -0.03 &  &  &  & 0.16 \\ 
  \citet{holtzman.2020} & 0.59 & 0.02 & 0.04 & 0.02 & 0.01 & 0 & -0.04 & 0.16 \\ 
   \hline
\end{tabular}
\caption{Fit fixed effect coefficients for each model along with the residual model variance. If only one model is compared to a baseline, there is a value for intercept and $\beta_{1}$. If more than one model, there is an additional parameter for each model. Because we use contrast coding, each coefficient can be interpreted as the difference from the grand mean.} 
\label{table:fixef}
\end{table*}

\begin{table*}[]
\centering \small
\begin{tabular}{lrrrrrrrr}
  \hline
Dataset & $\hat{\sigma}_{0w}$ & $\hat{\sigma}_{1w}$ & $\hat{\sigma}_{2w}$  & $\hat{\sigma}_{3w}$  & $\hat{\sigma}_{4w}$  & $\hat{\sigma}_{5w}$  & $\hat{\sigma}_{6w}$  \\ 
  \hline
\citet{hashimoto2019unifying} (LM) & 0 & 0.11 & 0.11 &  &  &  &  \\ 
  \citet{hashimoto2019unifying} (summarization) & 0 & 0.13 & 0.11 &  &  &  &  \\ 
    \citet{hashimoto2019unifying} (Reddit) & 0.11 & 0.04 & 0.08 & 0.06 & 0.17 &  &  \\ 
  WMT19 & 0.07 & 0.04 & 0.13 &  &  &  &  \\ 
  \citet{dathathri.2020}  & 0 & 0.04 & 0.05 & 0.05 & 0.05 &  &  \\ 
  \citet{holtzman.2020} & 0.09 & 0.05 & 0.03 & 0.04 & 0.04 & 0.02 & 0.04 \\ 
   \hline
\end{tabular}
\caption{Fit random effects standard deviations for worker. As in the equations above, $\hat{\sigma}_{0w}$ is the worker intercept and the rest of the parameters are worker slopes for each model.} 
\label{table:worker}
\end{table*}

\begin{table*}[]
\centering \small
\begin{tabular}{lrrrrrrrr}
  \hline
Dataset & $\hat{\sigma}_{0i}$ & $\hat{\sigma}_{1i}$ & $\hat{\sigma}_{2i}$  & $\hat{\sigma}_{3i}$  & $\hat{\sigma}_{4i}$  & $\hat{\sigma}_{5i}$  & $\hat{\sigma}_{6i}$  \\ 
  \hline
\citet{hashimoto2019unifying} (LM) & 0.04 & 0.14 & 0.1 &  &  &  &  \\ 
  \citet{hashimoto2019unifying} (summarization) & 0.07 & 0 & 0.18 &  &  &  &  \\
  \citet{hashimoto2019unifying} (Reddit) & 0 & 0.13 & 0.11 & 0.14 & 0.14 &  &  \\ 
  WMT19 & 0.05 & 0.03 & 0.15 &  &  &  &  \\ 
  \citet{dathathri.2020}  & 0 & 0.16 & 0.19 & 0.16 & 0.16 &  &  \\ 
  \citet{holtzman.2020} & 0 & 0.13 & 0.1 & 0.12 & 0.11 & 0.13 & 0.13 \\ 
   \hline
\end{tabular}
\caption{Fit random effects standard deviations for item. As in the equations above, $\hat{\sigma}_{0i}$ is the item intercept and the rest of the parameters are item slopes for each model.} 
\label{table:item}
\end{table*}

\begin{table*}[]
\centering \small
\begin{tabular}{lrrrrrrrr}
  \hline
Scenario & ${\sigma}_{w0}$ & ${\sigma}_{w1}$ & ${\sigma}_{i0}$ & ${\sigma}_{i1}$ &  ${\sigma}_{wi}$ \\ 
  \hline
Low variance & 0.01 & 0.04 & 0.01 & 0.13 & 0.16 \\ 
  High variance & 0.01 & 0.11 & 0.04 & 0.14 & 0.26 \\ 
     \hline
\end{tabular}
\caption{An example of high variance and low variance settings. The standard deviations correspond to the variance parameters for worker intercept, worker slope, item intercept, item slope, and sigma, respectively.} 
\label{table:lmer_recs}
\end{table*}

\subsection{Head to head human evaluations}

Another commonly used form of human evaluation is head to head comparison, where raters are shown a pair of outputs (one from each model), and asked to choose which they prefer, sometimes with ``neither'' as a third option. Head to head comparisons offer some advantages over ratings-basd approaches \citep{yannakakis.2015,van2019best}, but do not scale as well when comparing many models.

As with ordinal judgements, there are multiple ways of analyzing such data. If we treat annotator judgements as independent and identically distributed (such as if we only collect one judgement from each annotator), we could model this simply in terms of the underlying probabilities that a random annotator will prefer each model (as in the opening example in the main paper). In that case, running a power analysis would be a simple as assuming values for the underlying probabilities of each category (win, lose, draw), as usual based on pilot data or prior assumptions, and simulating many draws from that prior, checking in each sample to see if there is a statistically significant difference between win and lose. 

On the other hand, if multiple judgements will be collected from each annotator and/or for each pair of outputs, then it makes sense to use a richer model to account for all sources of variation, as described above (see \S\ref{app:lmer}). In particular, the mixed effects framework can be adopted, potentially by modeling the outcome as a logistic model (in the case of win or lose), with ties either excluded or split.

\clearpage
\onecolumn
\section{Additional Plots of Model Overlap}
\label{app:modeloverlapplots}

\begin{figure}[!ht]
    \centering
   \includegraphics[width=.71\linewidth]{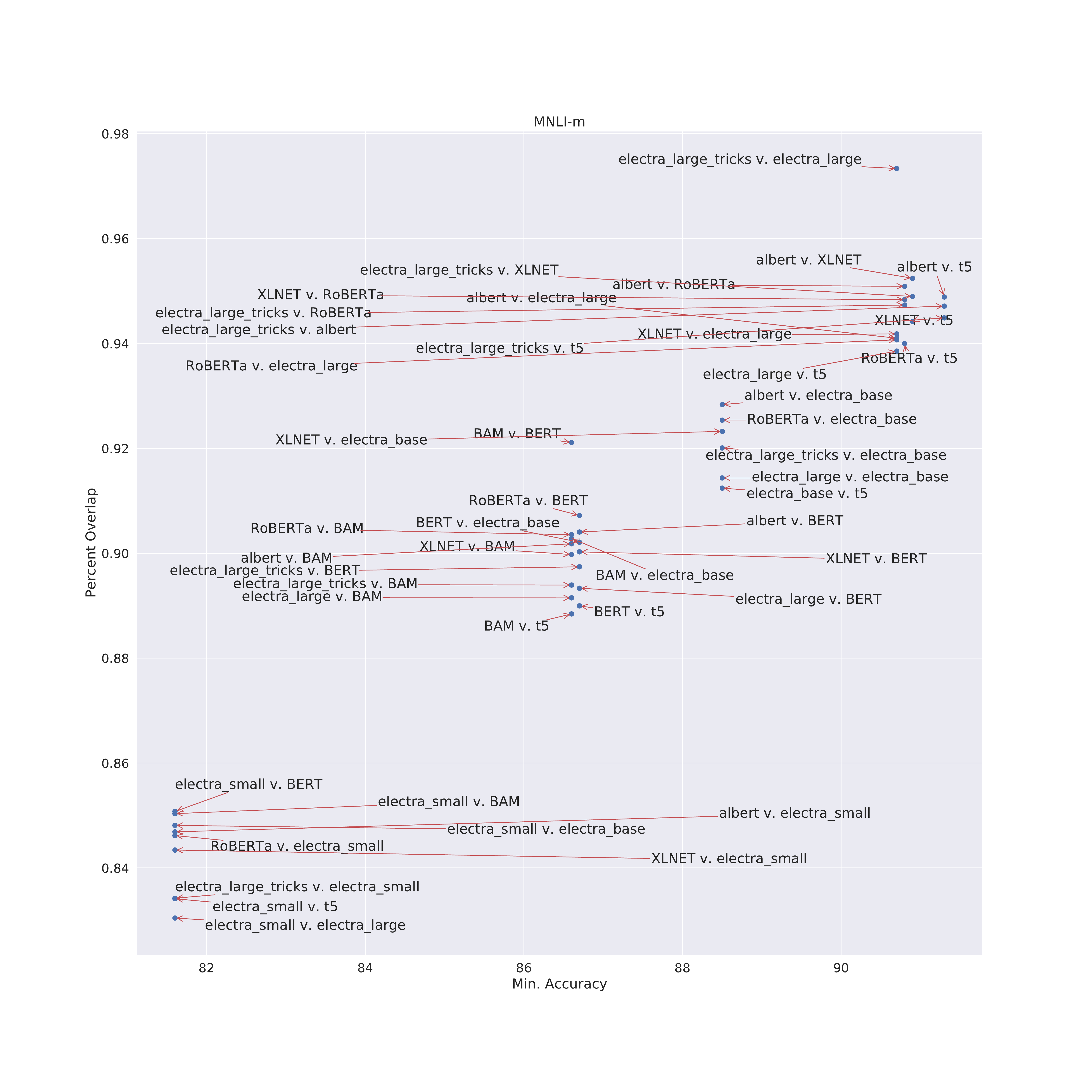}
   \includegraphics[width=.71\linewidth]{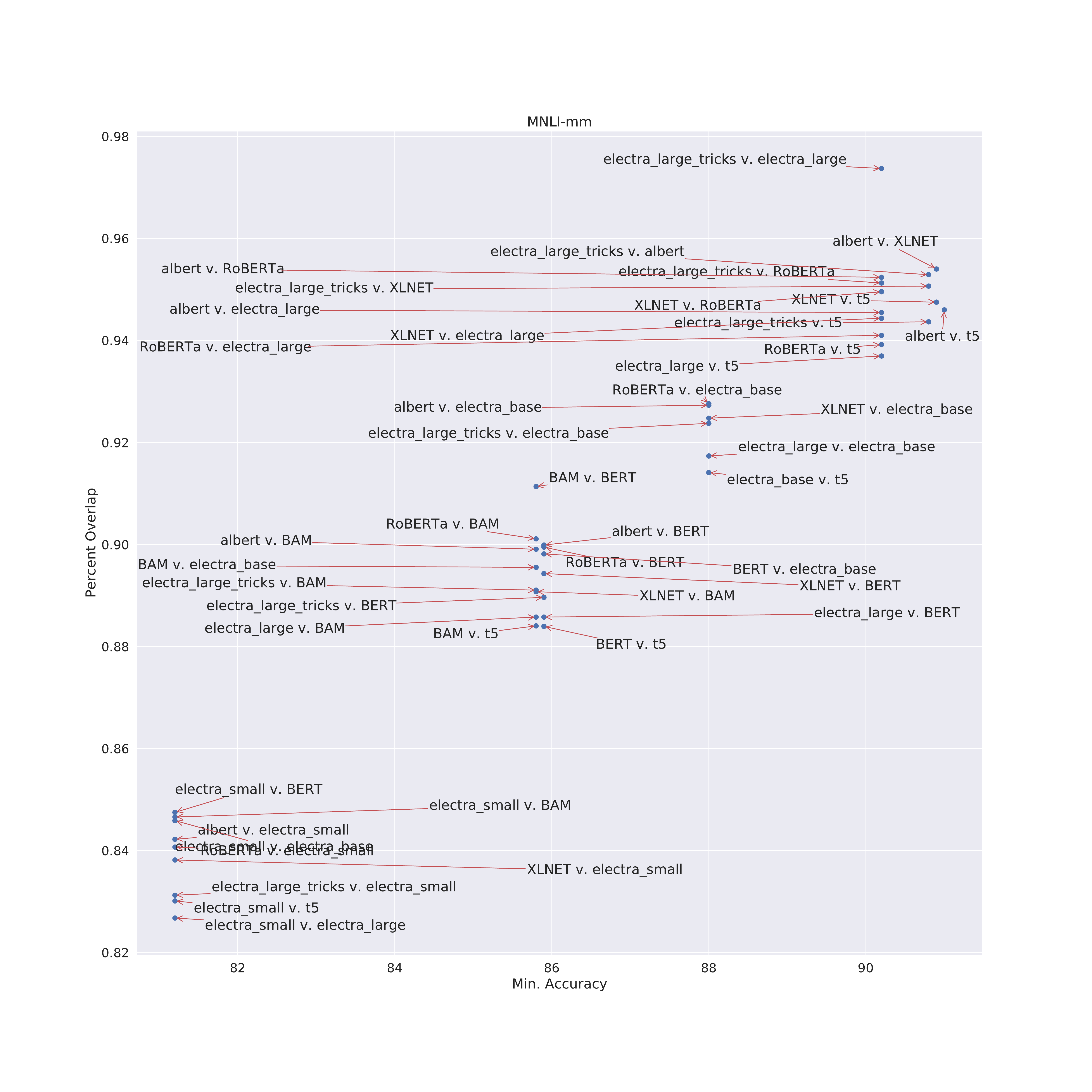}
    \label{fig:glueoverlapmnli}
\end{figure}

\begin{figure}
    \centering
      \includegraphics[width=.75\linewidth]{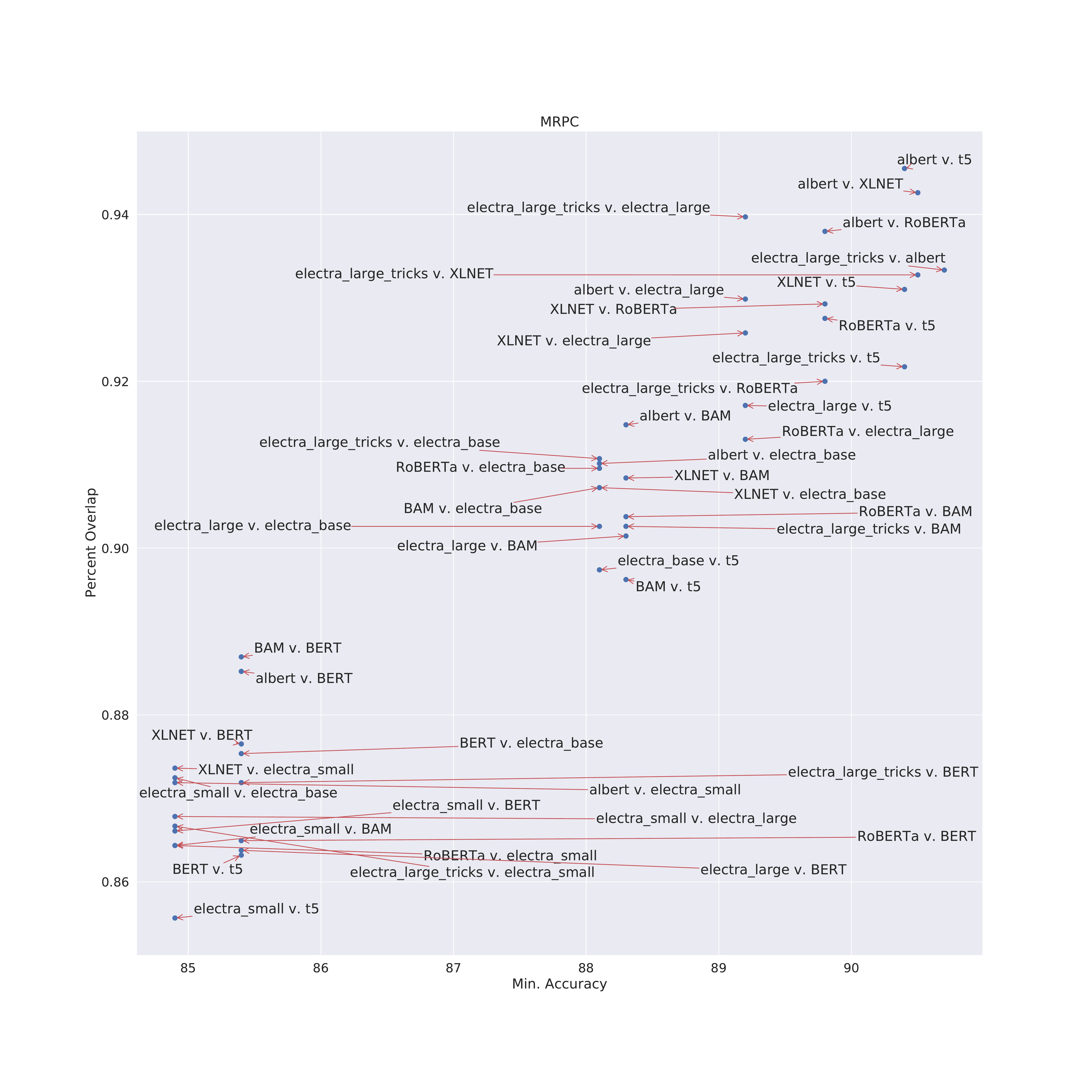}
      \includegraphics[width=.75\linewidth]{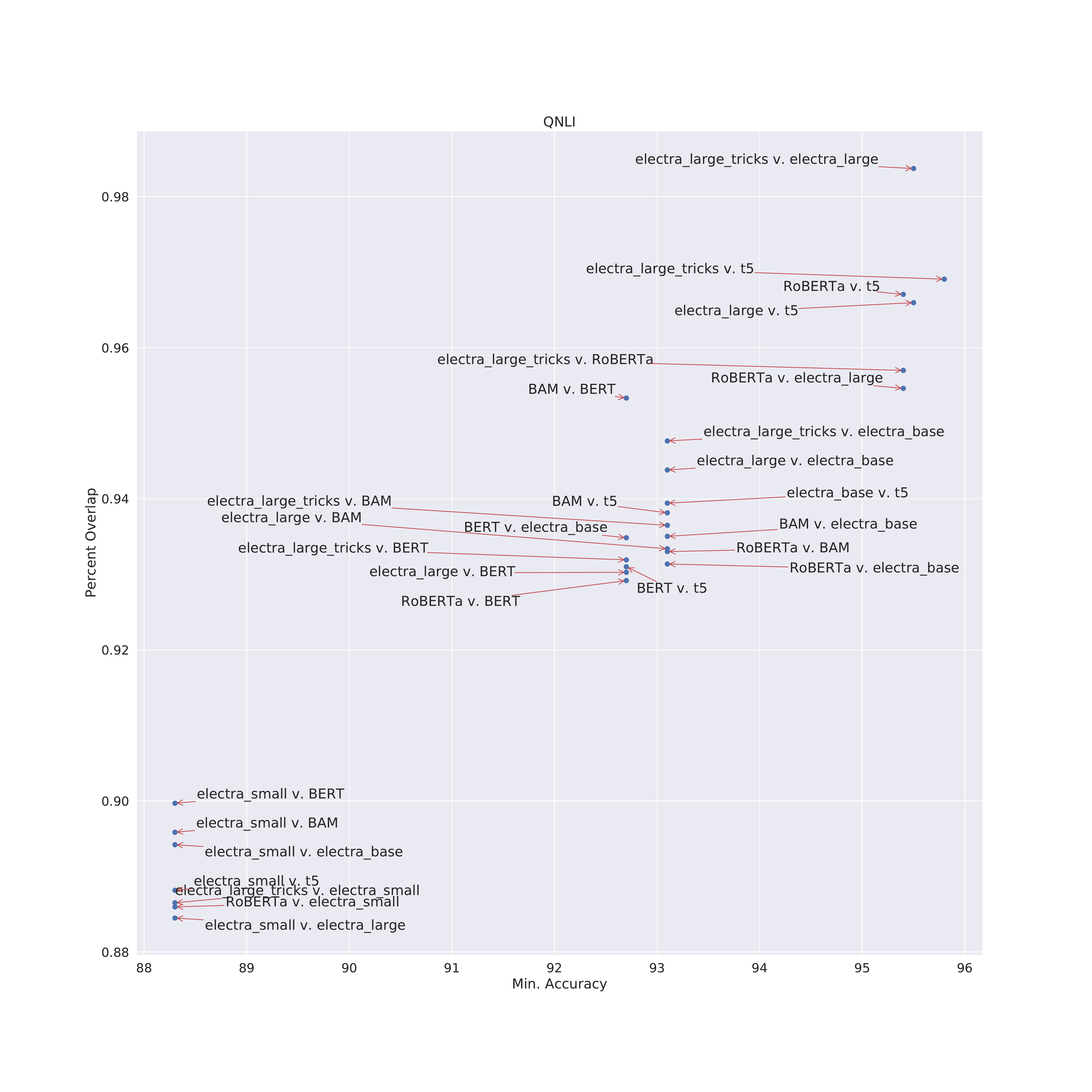}
\end{figure}

\begin{figure}
    \centering
      \includegraphics[width=.75\linewidth]{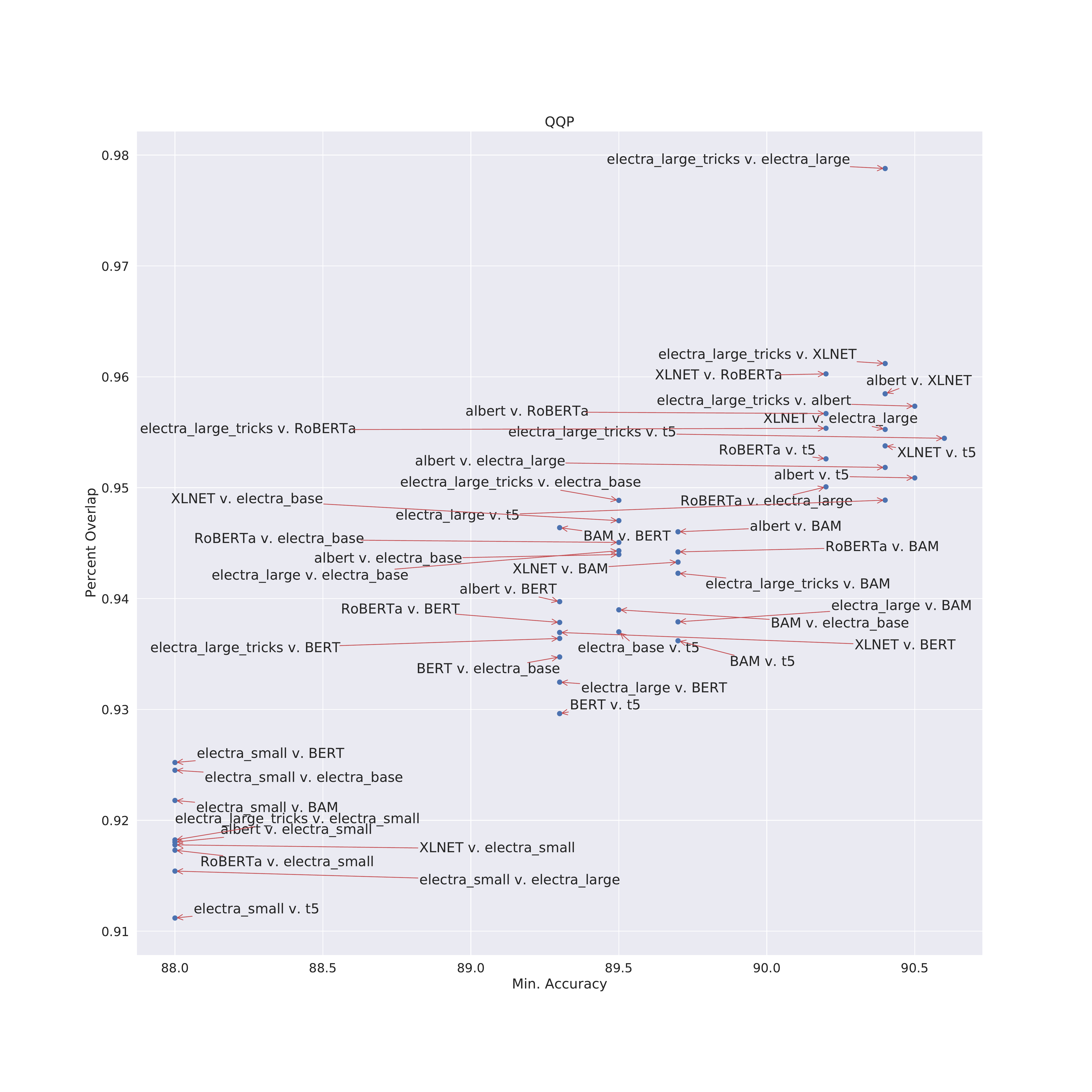}
            \includegraphics[width=.75\linewidth]{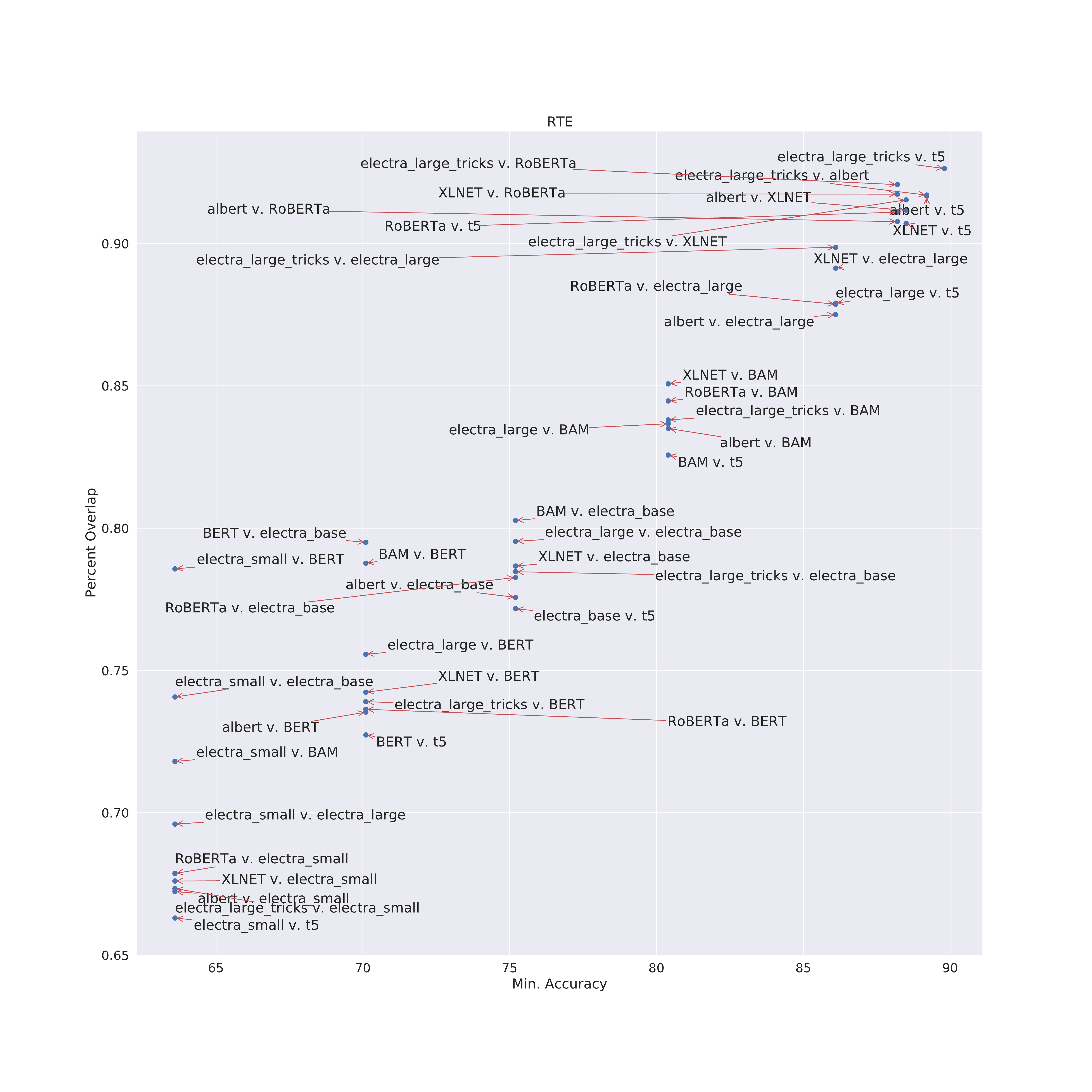}
\end{figure}

\begin{figure}
    \centering
      \includegraphics[width=.75\linewidth]{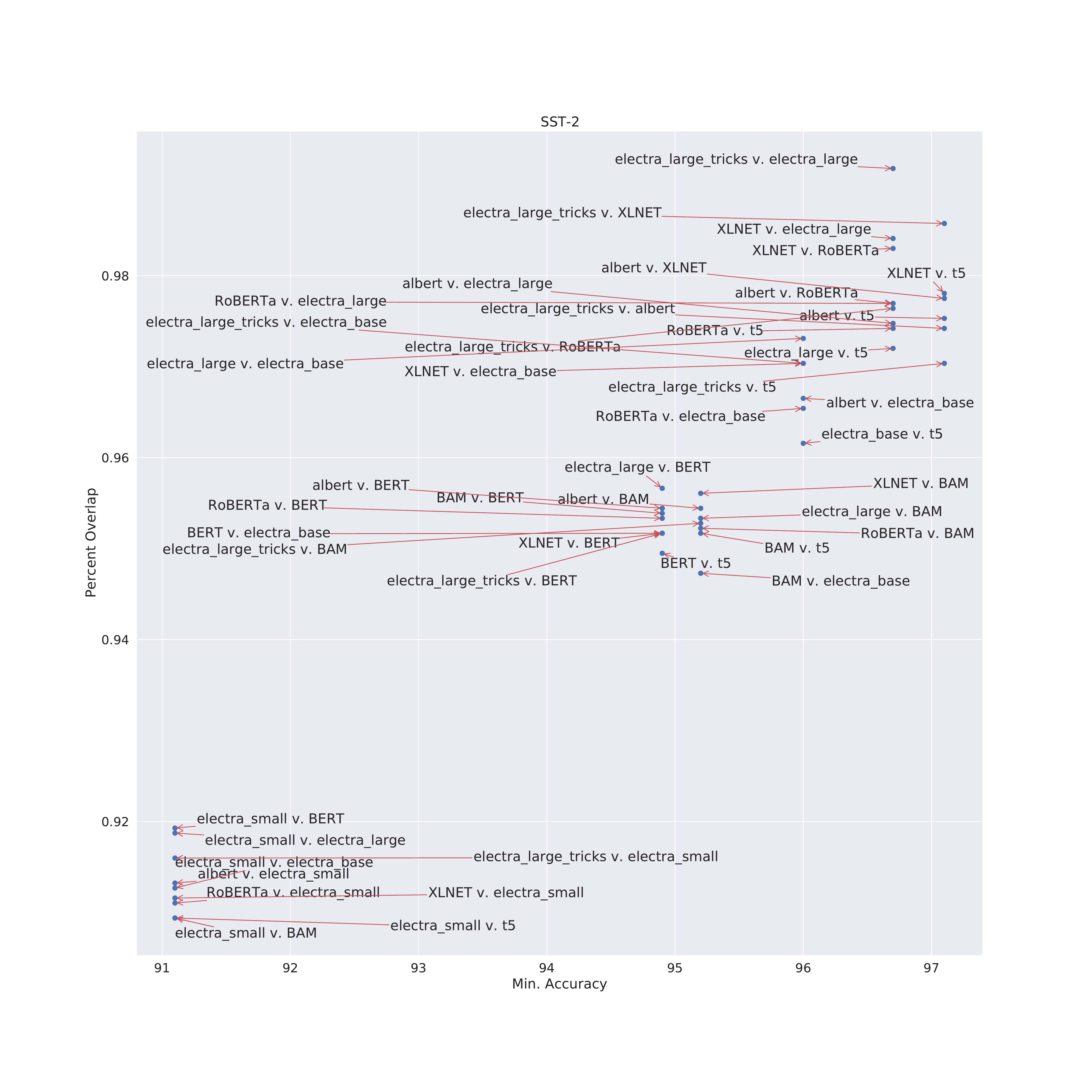}
      \includegraphics[width=.75\linewidth]{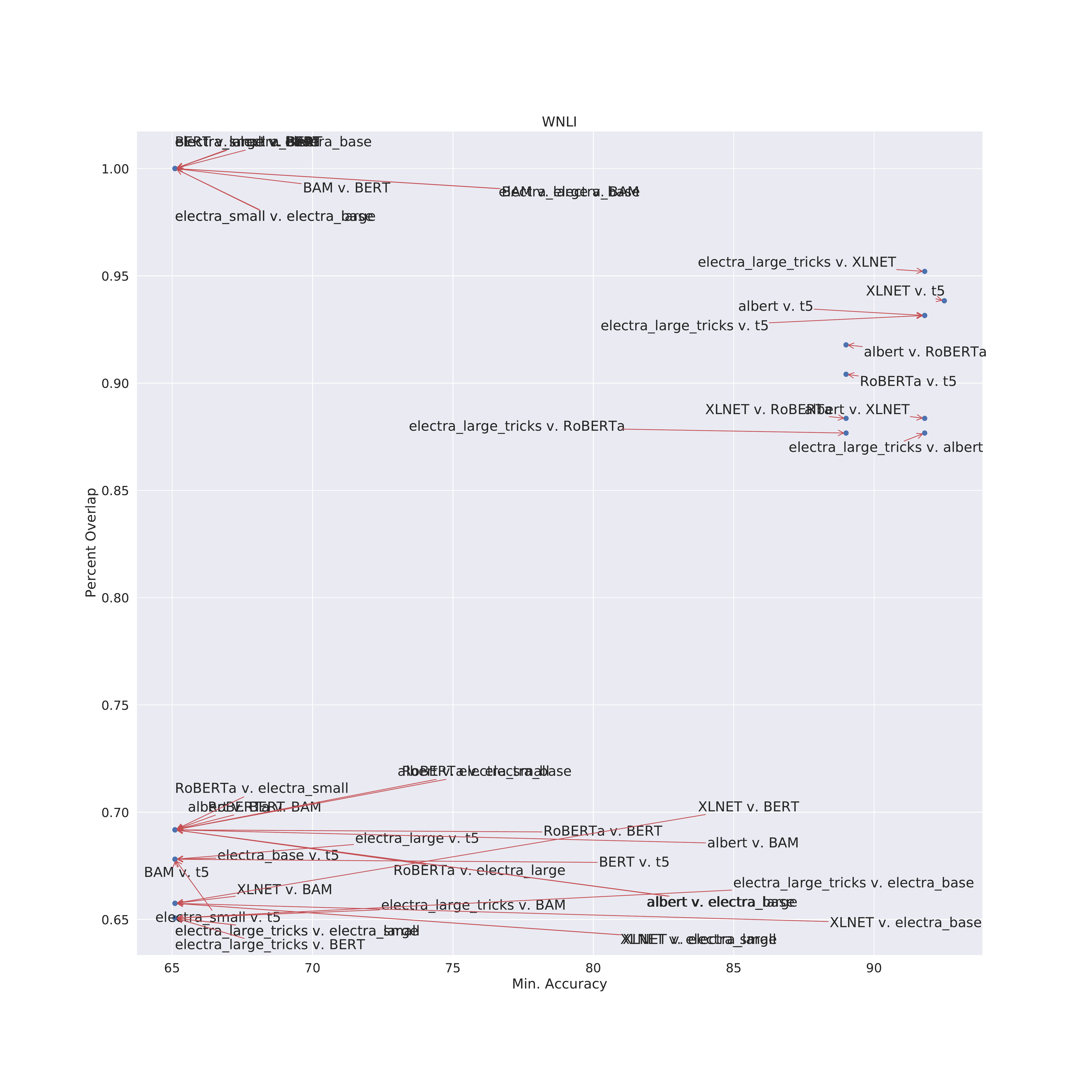}
\end{figure}
\end{document}